\documentclass{article}

\usepackage[preprint]{corl_2026} 

\usepackage{graphicx}
\usepackage{float}
\usepackage{xcolor}
\usepackage{booktabs}
\usepackage{graphicx}
\usepackage{caption}
\usepackage{amsfonts}
\usepackage{amsmath}
\usepackage{todonotes}
\usepackage{enumitem}
\usepackage{wrapfig}
\usepackage{etoc}
\usepackage{dsfont}

\definecolor{citecolor}{HTML}{0071BC}
\hypersetup{colorlinks,linkcolor={red},citecolor={citecolor}}

\usepackage{titlesec}
\titlespacing*{\section}{0pt}{6pt}{6pt}
\titlespacing*{\subsection}{0pt}{3.5pt}{3.5pt}

\title{CoorDex: Coordinating Body and Hand Priors for Continuous Dexterous Humanoid Loco-Manipulation}

\author{
    Sikai Li$^{1}$, Shuning Li$^{1}$, Zhenyu Wei$^{1}$, Yunchao Yao$^{1}$, Chenran Li$^{2}$, Mingyu Ding$^{1}$ \\
    $^1$University of North Carolina at Chapel Hill \quad $^2$University of California, Berkeley
}

\begin{document}
\maketitle

\vspace{-30pt}

\begin{figure}[H]
    \centering
    \includegraphics[width=\linewidth]{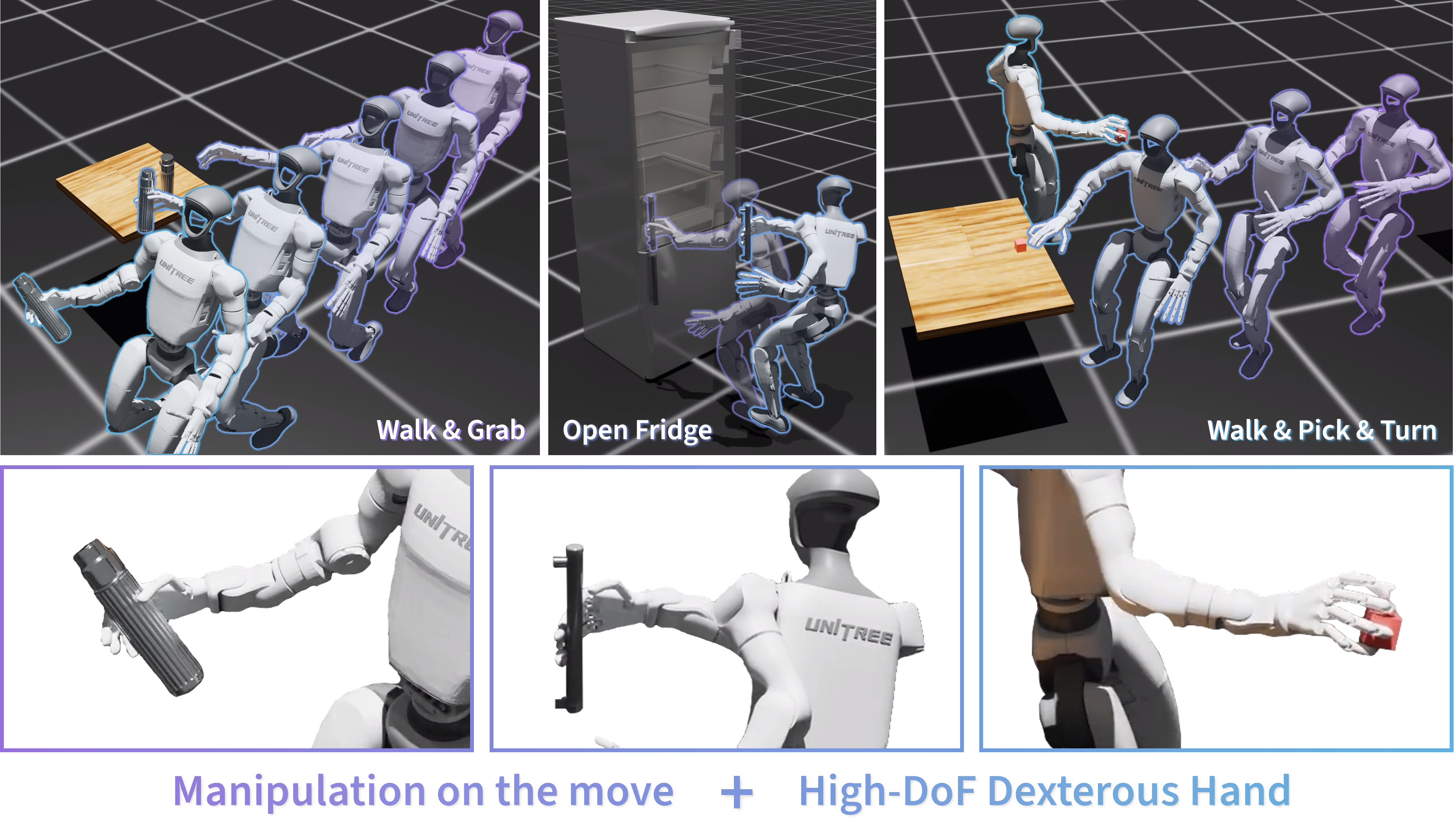}
    \caption{
    \textbf{Dexterous loco-manipulation on the move.}
    CoorDex enables a humanoid equipped with high-DoF dexterous hands to perform continuous loco-manipulation tasks that require simultaneous coordination between locomotion and dexterous hand control, such as walk-grasp-carry, fridge opening while stepping back, and walk-pick-turn.
    }
    \label{fig:teaser}
\end{figure}

\vspace{-17pt}

\begin{abstract}
    Humanoid loco-manipulation is often simplified into a stop-and-go process: walking to an object, stopping to manipulate it, and then resuming locomotion. It also commonly relies on low degree-of-freedom (DoF) end effectors that behave like an open-close grasp primitive. We introduce \textbf{CoorDex}, a learning pipeline that converts high-dimensional body and dexterous hand control into coordinated latent residual control, enabling high-DoF dexterous loco-manipulation on the move.
    Starting from simulated whole-body and hand demonstrations, CoorDex trains privileged motion tracking teachers for the humanoid body and dexterous hand, distills them into proprioception-conditioned latent priors, and uses the frozen priors as the action space for downstream residual reinforcement learning. A coordinated latent residual policy composes these priors through shared task context and separate body-hand residual heads, preserving natural whole-body motion while improving finger-level contact reliability.
    CoorDex enables a Unitree G1 humanoid with a 20-DoF WUJI hand to execute dexterous manipulation while in motion, including non-stop bottle grasping and carrying, fridge door opening on the move, and cube pick-and-turn. Ablations on the walk-grasp-carry task show that joint-space PPO, joint-space hand control, and monolithic latent prediction all fail under the same reward budget, while the latent-prior interface and coordinated residual structure make high-dimensional contact-rich loco-manipulation trainable. Project Page: \url{https://skevinci.github.io/coordex/} 
\end{abstract}



\section{Introduction}

Humanoid robots equipped with high-DoF dexterous hands promise a single embodiment that can both navigate human environments and manipulate everyday objects. Realizing this promise requires more than placing an end effector near a target. 
The robot must maintain balance, keep the object reachable, coordinate wrist and finger poses before contact, close the hand without disturbing the object, and continue transporting the object while hand-object interactions perturb whole-body dynamics.
Learning such behaviors directly over all actuated joints creates a challenging high-dimensional exploration problem, where locomotion and dexterous hand control must be simultaneously coordinated throughout execution.

Recent progress has advanced complementary parts of humanoid loco-manipulation. Whole-body tracking enables stable locomotion and expressive upper-body motion~\cite{Peng_2018, liao2025beyondmimicmotiontrackingversatile, luo2026sonicsupersizingmotiontracking, he2024learninghumantohumanoidrealtimewholebody, cheng2024expressivewholebodycontrolhumanoid, ji2025exbody2advancedexpressivehumanoid, he2025hoverversatileneuralwholebody}. Physics-based priors reduce exploration through structured latent action spaces~\cite{Peng_2021, Peng_2022, tessler2023calm, luo2024pulse, tan2026sphericallatentmotionprior}. Teleoperation, residual learning, hierarchical control, force-adaptive policies, and vision-conditioned systems further improve task-level interaction and sim-to-real transfer~\cite{he2024omnih2ouniversaldexteroushumantohumanoid, fu2024humanplushumanoidshadowingimitation, heng2026humdex, zhao2025resmimic, fu2025demohlmdemonstrationgeneralizablehumanoid, kuang2025skillblenderversatilehumanoidwholebody, zhang2025falconlearningforceadaptivehumanoid, sun2025ulc, viral2025, xue2025openingsimtorealdoorhumanoid, jiang2025wholebodyvlaunifiedlatentvla}. However, many systems still use end-effector-centric, low-dimensional, or primitive-based manipulation interfaces, leaving high-DoF multi-finger contact during continuous locomotion remains less isolated and less understood.

This limitation is especially pronounced in on-the-move dexterous loco-manipulation. 
Current hand-centric methods have made strong progress in grasp synthesis, hand-object representation, cross-embodiment transfer, and demonstration generation~\cite{wang2023dexgraspnetlargescaleroboticdexterous, li2023gendexgraspgeneralizabledexterousgrasping, zhan2024oakink2datasetbimanualhandsobject, wei2025mathcaldrograspunifiedrepresentation, wei2026handruleallcanonical, li2025maniptransefficientdexterousbimanual, jiang2025dexmimicgenautomateddatageneration, brüdigam2024jactaversatileplannerlearning}. Yet they often assume that the wrist or arm trajectory is provided by teleoperation, planning, a fixed-base setup, or a stationary whole-body controller.
As a result, existing humanoid loco-manipulation systems are often simplified into a stop-and-go process: walking to an object, stopping to manipulate it, and then resuming locomotion.
On a walking humanoid, however, the wrist pose emerges from foot timing, root motion, torso posture, and whole-body reaching. A hand prior that must also explain 6D wrist motion therefore spends much of its capacity on body-side placement rather than on finger coordination. The key challenge is to decompose and coordinate body-side wrist placement with hand-side finger motion during continuous locomotion.

We introduce \textbf{CoorDex}, a learning pipeline that makes high-DoF dexterous humanoid loco-manipulation trainable through coordinated latent residual control.
It targets dexterous manipulation on the move, where wrist placement emerges from whole-body motion, by separating whole-body wrist placement from finger coordination, modeling them with a body prior and a wrist-stabilized hand prior, and coupling them through a coordinated latent residual policy.
%
Specifically, CoorDex first constructs separate latent priors for the humanoid body and dexterous hand. Each prior is trained from a privileged motion tracking teacher and distilled into a proprioception conditioned student prior and decoder using a variational bottleneck~\cite{luo2024pulse}.
The body prior supports locomotion, reaching, and wrist placement, while the wrist-stabilized hand prior controls only the active finger joints and captures reusable finger coordination. This factorization replaces direct full-joint control with residual control over two compact latent spaces, separating body-side placement from hand-side dexterity while preserving their downstream coordination.

To compose these priors, CoorDex uses a coordinated latent residual policy. At each control step, the frozen body and hand priors first encode their respective proprioception into latent means. These means, together with object-relative geometry, contact features, and the current proprioceptive state, are passed to a shared coordination trunk to produce a task-level coordination feature. Two residual heads then predict body and hand latent corrections, which are added to the corresponding prior means before decoding. The body residual adjusts stepping, torso motion, reaching, and wrist placement, whereas the hand residual refines finger preshape, closure, and contact. This design couples the two subsystems through shared task state without collapsing them into a single monolithic action head. It preserves the structural separation between whole-body motion and finger-level dexterity.



Our contributions are threefold.
 \textbf{1)} We present a complete learning pipeline in Isaac Lab~\cite{mittal2025isaaclab} for high-DoF dexterous humanoid loco-manipulation, including simulated whole-body and hand demonstration collection, motion tracking policies, prior distillation, and downstream residual RL.
 \textbf{2)} We propose an asymmetric body-hand prior composition for high-dimensional control. Wrist placement emerges from whole-body motion from a task-aligned body prior, while a wrist-stabilized hand prior captures reusable finger skills. A coordinated latent residual policy adapts the frozen priors through shared task context and body-hand residual heads.
 \textbf{3)} We provide ablations on the walk-grasp-carry task showing that raw joint-space PPO, the body prior with joint-space hand control, and monolithic latent prediction all fail under the same reward budget, highlighting the importance of both latent-prior actions and structured body-hand coordination for dexterous loco-manipulation.



\section{Related Work}
\textbf{Motion Imitation and Motion Priors.}
Physics-based motion imitation enables high-dimensional humanoid control. DeepMimic~\cite{Peng_2018} combines reference motion tracking with task rewards to produce robust simulated skills. Recent humanoid trackers scale this idea to full-body robots by improving motion retargeting, adaptive tracking, privileged training, and sim-to-real deployment, enabling expressive locomotion, whole-body teleoperation, and highly dynamic behaviors~\cite{luo2026sonicsupersizingmotiontracking, he2024learninghumantohumanoidrealtimewholebody, cheng2024expressivewholebodycontrolhumanoid, ji2025exbody2advancedexpressivehumanoid, liao2025beyondmimicmotiontrackingversatile, xie2025kungfubotphysicsbasedhumanoidwholebody, he2025hoverversatileneuralwholebody, ze2025twist2scalableportableholistic, ze2025twistteleoperatedwholebodyimitation, li2025amoadaptivemotionoptimization, chen2025gmtgeneralmotiontracking}. A related line learns latent motion priors that replace raw joint actions with more structured action spaces. AMP~\cite{Peng_2021} and ASE~\cite{Peng_2022} use adversarial objectives to regularize or parameterize downstream control, while CALM~\cite{tessler2023calm}, PULSE~\cite{luo2024pulse}, and SLMP~\cite{tan2026sphericallatentmotionprior} further study conditional or compact latent representations for reusable physics-based skills. Building on prior work that learns motion priors for humanoid body control, we extend the same principle to dexterous hand control by constructing a separate hand prior. CoorDex then studies how to compose the body and hand priors for downstream contact-rich loco-manipulation tasks.

\textbf{Dexterous Hand Manipulation.}
Dexterous hand research provides tools for grasp synthesis, hand-object representation, and manipulation data generation. Large-scale datasets and generative methods synthesize diverse multi-finger grasps over objects and hand morphologies~\cite{wang2023dexgraspnetlargescaleroboticdexterous, li2023gendexgraspgeneralizabledexterousgrasping, zhan2024oakink2datasetbimanualhandsobject,zhao2025dexh2r,zhang2026unidex,liang2025dexhanddiff}. Cross-embodiment representations such as $\mathcal{D}(\mathcal{R}, \mathcal{O})$ Grasp~\cite{wei2025mathcaldrograspunifiedrepresentation} and OHRA~\cite{wei2026handruleallcanonical} encode hand-object geometry or hand morphology in forms that transfer across different dexterous hands. Demonstration generation and transfer systems such as ManipTrans~\cite{li2025maniptransefficientdexterousbimanual}, DexMimicGen~\cite{jiang2025dexmimicgenautomateddatageneration}, and Jacta~\cite{brüdigam2024jactaversatileplannerlearning} use human motions, planners, or few demonstrations to bootstrap dexterous manipulation through tracking, imitation, and residual refinement. 
While these methods provide effective supervision for finger-level dexterity, they typically assume that wrist motion is externally provided. CoorDex targets a walking humanoid, where wrist placement emerges from whole-body motion. 

\textbf{Humanoid Loco-Manipulation.}
Humanoid loco-manipulation couples locomotion, reaching, contact, and object transport. Existing approaches address this coupling from several directions. Optimization and residual-learning methods generate or refine dynamically feasible whole-body behaviors for object interaction~\cite{liu2025opt2skillimitatingdynamicallyfeasiblewholebody, zhao2025resmimic, fu2025demohlmdemonstrationgeneralizablehumanoid}. Skill composition and unified whole-body controllers widen the range of reachable tasks, while force-adaptive policies make humanoids more robust to payloads and contacts~\cite{kuang2025skillblenderversatilehumanoidwholebody, sun2025ulc, zhang2025falconlearningforceadaptivehumanoid}. Teleoperation systems provide high-quality humanoid demonstrations and have begun to include dexterous hands~\cite{he2024omnih2ouniversaldexteroushumantohumanoid, homie2025, heng2026humdex}. Vision-based systems push loco-manipulation toward large-scale sim-to-real transfer and language- or video-conditioned control~\cite{viral2025, xue2025openingsimtorealdoorhumanoid, jiang2025wholebodyvlaunifiedlatentvla}. We isolate the control problem of continuous high-DoF dexterous loco-manipulation, where grasp formation and object transport occur during ongoing walking rather than in a stationary manipulation phase. This setting is closer to finger-level humanoid dexterity than to standard mobile manipulation, where navigation and manipulation can be separated in time.

\section{Method}

\begin{figure}[t]
    \centering
    \includegraphics[width=\linewidth]{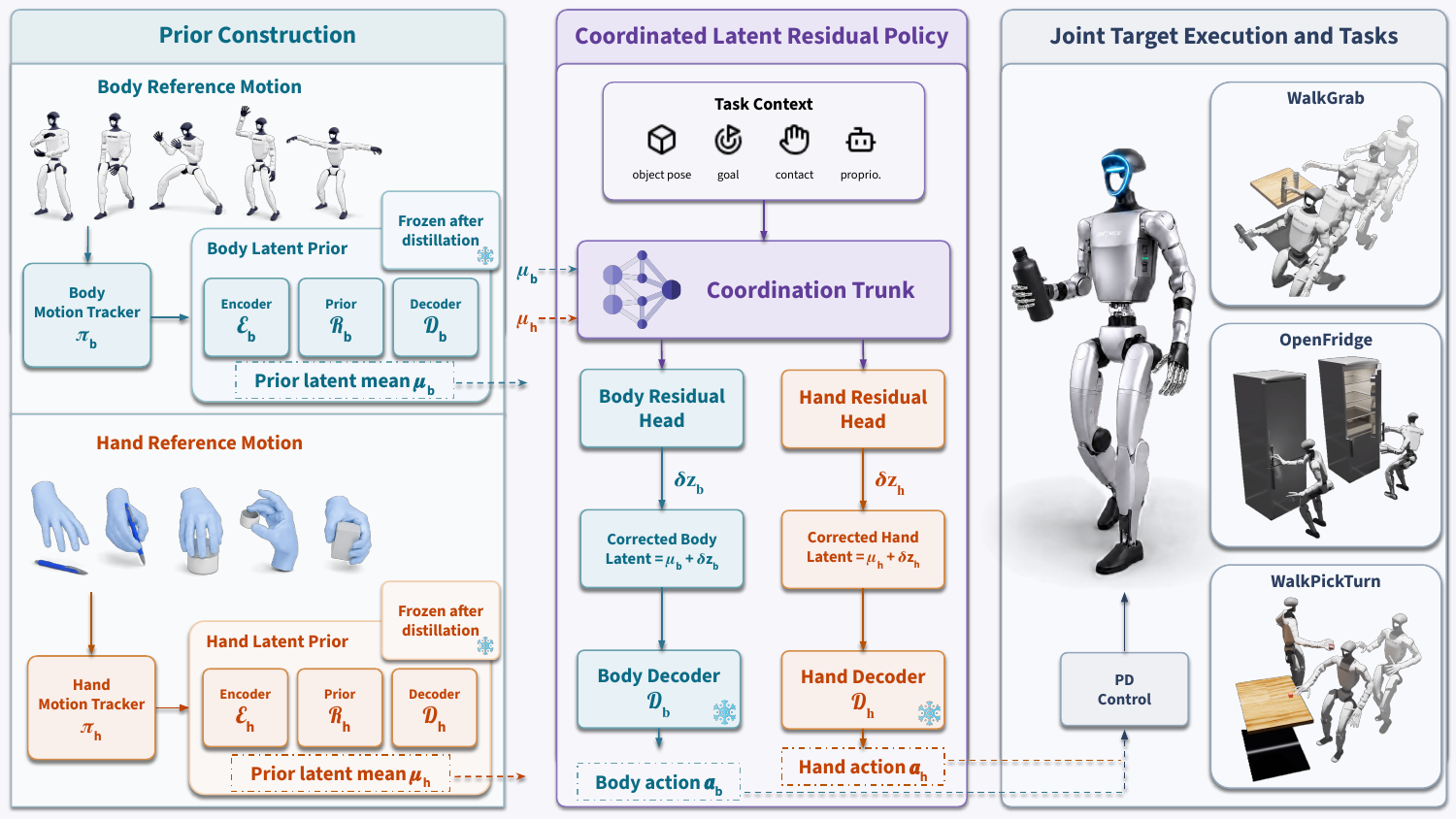}
    \vspace{-14pt}
    \caption{
        \textbf{Overview of CoorDex}. Body and hand reference motions are tracked by privileged teachers and distilled into separate proprioception-conditioned latent priors. During downstream RL, a coordinated residual policy uses task context and prior means to predict body and hand latent residuals. The frozen decoders map the corrected latents to joint-position targets for loco-manipulation.
    }
    \vspace{-8pt}
    \label{fig:method}
\end{figure}

We present CoorDex, a modular pipeline that maps high-dimensional humanoid locomotion and dexterous hand control to coordinated latent residual control (Fig.~\ref{fig:method}). It builds separate body and hand priors and trains a downstream residual RL policy to coordinate them. Sec.~\ref{sec:prior_construction} describes prior construction via teacher tracking and proprioceptive distillation; Sec.~\ref{sec:coord_policy} introduces the coordinated latent residual policy; and Sec.~\ref{sec:task_design} details residual PPO~\cite{schulman2017proximalpolicyoptimizationalgorithms} training and environment design.

\subsection{Prior Construction}
\label{sec:prior_construction}

We construct two separate motion priors, one for the humanoid body and one for the dexterous hand. The two priors are trained on different tracking tasks but share the same distillation pipeline. Let $x \in \{b,h\}$ denote the body or hand subsystem.

\textbf{Demonstration collection in simulation}
We collect task reference motions in Isaac Lab through a simulated teleoperation pipeline. Lower-body locomotion is generated by an AGILE-based locomotion controller~\cite{zhao2026agilecomprehensiveworkflowhumanoid}, while the operator provides right wrist and hand motion through Apple Vision Pro using the CloudXR-based XR teleoperation interface in Isaac Lab~\cite{mittal2025isaaclab}. The tracked human wrist pose serves as the end-effector target for a Pink inverse-kinematics solver, and human hand motion is retargeted to the target dexterous hand with optimization-based dex-retargeting~\cite{qin2023anyteleop}. The resulting whole-body trajectories provide the reference motions for body tracking training.

\textbf{Humanoid body prior.}
For the body subsystem, we first train a general whole-body motion tracking teacher $\pi_T^b$~\cite{liao2025beyondmimicmotiontrackingversatile}. 
The teacher takes body proprioception $\mathbf{s}^{b,p}_t$ and a reference goal $\mathbf{s}^{b,g}_t$ as input, and outputs joint position targets $\mathbf{a}^{b,T}_t$ for the body joints.
Before training, we preprocess each reference motion on the humanoid model equipped with the corresponding dexterous hands. The resulting reference trajectories thus match the morphology and kinematic structure used in our downstream loco-manipulation
tasks.

\textbf{Dexterous hand prior.}
For the hand subsystem, we train a privileged hand tracking teacher $\pi_T^h$ in a floating-hand environment using ManipTrans-style~\cite{li2025maniptransefficientdexterousbimanual} retargeted hand-object motions. 
The teacher observes hand proprioception $\mathbf{s}^{h,p}_t$ and a reference goal $\mathbf{s}^{h,g}_t$, and outputs joint position targets $\mathbf{a}^{h,T}_t$ for the active finger joints. During hand-prior training, the reference wrist pose and velocity are written directly into simulation, so the teacher and the learned prior control only finger motion. This wrist-stabilized design keeps the hand latent space from spending most of its capacity on 6D wrist motion, and makes the learned latent command directly useful for finger coordination.

\textbf{Distillation.}
After training the tracking teachers, we distill each teacher into a proprioception conditioned latent prior~\cite{luo2024pulse}. For each subsystem $x \in \{b,h\}$, the student contains an encoder $\mathcal{E}_x(\mathbf{z}^x_t \mid \mathbf{s}^{x,g}_t,\mathbf{s}^{x,p}_t)$, a proprioceptive prior $\mathcal{R}_x(\mathbf{z}^x_t \mid \mathbf{s}^{x,p}_t)$, and a decoder $D_x(\mathbf{s}^{x,p}_t,\mathbf{z}^x_t)$. The training objective combines teacher-action reconstruction, temporal smoothness of the encoder means, and KL regularization between the encoder and the prior:
\begin{equation}
    \mathcal{L}^{x}_{\mathrm{distill}}
    =
    \mathcal{L}^{x}_{\mathrm{action}}
    +
    \alpha_x \mathcal{L}^{x}_{\mathrm{regu}}
    +
    \beta_x \mathcal{L}^{x}_{\mathrm{KL}} .
\end{equation}
After distillation, $\mathcal{R}_x$ and $D_x$ are frozen, and the prior mean $\boldsymbol{\mu}^{x,p}_t$ provides the default latent command for downstream residual RL. Full definitions are given in Appendix~\ref{app:motion_prior_implementation}.

\subsection{Coordinated Latent Residual Policy}
\label{sec:coord_policy}

Given the frozen body and hand priors, a downstream residual PPO~\cite{schulman2017proximalpolicyoptimizationalgorithms} policy coordinates and adapts them for the loco-manipulation task. At each control step, we obtain the proprioception conditioned prior means from the frozen prior networks:
\begin{equation}
\small
    \boldsymbol{\mu}^{b,p}_t
    =
    \mathrm{Mean}
    \left[
    \mathcal{R}_{b}
    \left(
    \mathbf{z}^b_t
    \mid
    \mathbf{s}^{b,p}_t
    \right)
    \right],
    \qquad
    \boldsymbol{\mu}^{h,p}_t
    =
    \mathrm{Mean}
    \left[
    \mathcal{R}_{h}
    \left(
    \mathbf{z}^h_t
    \mid
    \mathbf{s}^{h,p}_t
    \right)
    \right].
\end{equation}
The actor predicts residuals in the two latent spaces rather than joint-space targets:
\begin{equation}
\small
    \Delta \mathbf{z}_t
    =
    \left[
    \Delta \mathbf{z}^{b}_t,
    \Delta \mathbf{z}^{h}_t
    \right],
    \qquad
    \Delta \mathbf{z}^{b}_t \in \mathbb{R}^{d_b},
    \quad
    \Delta \mathbf{z}^{h}_t \in \mathbb{R}^{d_h}.
    \label{eq:latent_residual}
\end{equation}

The actor first computes a shared coordination representation $\mathbf{c}_t$ to capture body-hand coupling, and then predicts residual commands through a specialized body head $f_b$ and hand head $f_h$:
\begin{equation}
\small
\begin{gathered}
\mathbf{c}_t =
f_{\mathrm{coord}}
\left(
\mathbf{s}^{b,p}_t,
\mathbf{s}^{h,p}_t,
\mathbf{s}^{\mathrm{task}}_t,
\mathbf{s}^{\mathrm{hand\text{-}object}}_t,
\boldsymbol{\mu}^{b,p}_t,
\boldsymbol{\mu}^{h,p}_t,
\Delta \mathbf{z}_{t-1}
\right), \\[3pt]
\Delta \mathbf{z}^{b}_t =
\tanh
\left(
f_b(\mathbf{c}_t, \mathbf{s}^{b,p}_t, \boldsymbol{\mu}^{b,p}_t)
\right),
\quad
\Delta \mathbf{z}^{h}_t =
\tanh
\left(
f_h(\mathbf{c}_t, \mathbf{s}^{h,p}_t, \boldsymbol{\mu}^{h,p}_t,
\mathbf{s}^{\mathrm{hand\text{-}object}}_t)
\right).
\end{gathered}
\label{eq:coord_policy}
\end{equation}
Here $\mathbf{s}^{\mathrm{task}}_t$ contains task-level state such as object pose, goal information, projected gravity, and contact features. The hand-object state $\mathbf{s}^{\mathrm{hand\text{-}object}}_t$ includes the object pose in the hand frame and fingertip-object contact features.

The corrected latent commands are
\begin{equation}
\small
    \tilde{\mathbf{z}}^{b}_t
    =
    \boldsymbol{\mu}^{b,p}_t
    +
    \Delta \mathbf{z}^{b}_t,
    \qquad
    \tilde{\mathbf{z}}^{h}_t
    =
    \boldsymbol{\mu}^{h,p}_t
    +
    \Delta \mathbf{z}^{h}_t.
\end{equation}
The final joint position targets are produced by the frozen decoders:
\begin{equation}
\small
    \mathbf{a}^{b}_t
    =
    D_b
    \left(
    \mathbf{s}^{b,p}_t,
    \tilde{\mathbf{z}}^{b}_t
    \right),
    \qquad
    \mathbf{a}^{h}_t
    =
    D_h
    \left(
    \mathbf{s}^{h,p}_t,
    \tilde{\mathbf{z}}^{h}_t
    \right).
    \label{eq:decode_action}
\end{equation}
The body decoder outputs targets for all humanoid body joints. The hand decoder outputs targets for the active finger joints of the selected dexterous hand. These targets are inserted into the corresponding joint slots and executed by a low-level PD controller.
This architecture keeps downstream exploration low-dimensional while preserving separate control authority for locomotion and grasping. The shared coordination trunk lets the policy reason about task phase and contact state. The separate heads let the body residual adapt stepping, torso motion, reaching, and wrist placement, and let the hand residual adapt finger preshape, closure, and contact refinement.

\subsection{Residual RL and Environment Design}
\label{sec:task_design}
We train downstream policies with PPO while keeping the body and hand priors frozen. The policy observation contains body proprioception, hand proprioception, task state, object-relative geometry, fingertip contact features, the prior means, and the previous latent residual. Since the actor outputs only $\Delta \mathbf{z}_t$, exploration happens in the learned body and hand latent spaces rather than in the full joint-position action space.

All tasks use the same latent residual interface but differ in their interaction structure and exploration difficulty. The priors used by different tasks are not identical, since each task needs different coarse motion support. For shorter interaction tasks such as walk-grasp-carry and fridge door opening, we use a single-stage reward without reference state initialization. The reward combines locomotion and balance terms, palm-object reaching and alignment terms, fingertip contact and sustained contact terms, task-completion terms such as lift, carry, or door-angle progress, and regularization on latent residuals, decoded action changes, and joint velocities. For the longer-horizon walk-pick-turn task, pure reward optimization from the initial state is unreliable. We therefore use a lightweight staged reward together with demonstration-free reference state initialization (NoDemoRSI), which resets part of the episodes from states the policy itself has reached. This curriculum is task-specific and is not part of the policy architecture. More details are given in Appendix~\ref{sec:impl}.

\section{Experimental Results}
\label{sec:experiments}

We evaluate the core design components of \textbf{CoorDex} in Isaac Lab~\cite{mittal2025isaaclab}. Our experiments use a 29-DoF Unitree G1 humanoid~\cite{unitree_g1} equipped with a 20-DoF five-finger WUJI dexterous hand~\cite{wuji_hand}. 
We study three questions. \textbf{Q1.} Whether a unified body--hand latent residual interface supports diverse loco-manipulation skills. \textbf{Q2.} Whether latent actions improve exploration over direct joint-space control. \textbf{Q3.} Whether coordinated residual prediction improves over monolithic latent prediction.

\subsection{Experiment Setup}
\label{sec:exp_setup}

\begin{wraptable}{r}{0.32\linewidth}
    \centering
    \vspace{-12pt}
    \caption{Actuated degrees of freedom and latent dimensions.}
    \vspace{-6pt}
    \label{tab:latent_dims}
    \small
    \setlength{\tabcolsep}{4pt}
    \begin{tabular}{lcc}
        \toprule
        Subsystem & DoF & Latent dim. \\
        \midrule
        G1 body & 29 & 16 \\
        WUJI hand & 20 & 12 \\
        \bottomrule
    \end{tabular}
    \vspace{-12pt}
\end{wraptable}

\textbf{Tasks.}
We consider three dexterous humanoid loco-manipulation tasks with different interaction structures. In \textsc{WalkGrab}, the humanoid must grasp and lift a bottle from a side table while continuously walking forward. In \textsc{OpenFridge}, it must grasp a fridge handle and open the door while stepping backward to create workspace. In \textsc{WalkPickTurn}, it must approach a table, pick up a cube, and complete a $180^\circ$ turn while retaining the object. The three tasks respectively stress dynamic grasping during locomotion, sustained articulated-object interaction, and long-horizon skill composition. Fig.~\ref{fig:teaser} visualizes the three tasks. Table~\ref{tab:latent_dims} summarizes the action dimensions.

\textbf{Evaluation metrics.}
Unless stated otherwise, all metrics are computed over $50{,}000$ evaluation episodes collected with $10{,}000$ parallel simulation environments. We report task success, fall rate, drop rate when applicable, and task-specific progress metrics. A rollout counts as successful only if the task-specific completion condition is satisfied and the robot stays balanced. \textsc{WalkGrab} requires lifting and carrying the bottle beyond a target displacement, \textsc{OpenFridge} requires the maximum door angle to exceed the success threshold, and \textsc{WalkPickTurn} requires grasping, lifting, and retaining the cube while completing a $180^\circ$ turn. Exact thresholds are provided in Appendix~\ref{tab:task_success}. We additionally report diagnostic metrics, including door angle, heading error, and action rate.

\textbf{Controlled variants on \textsc{WalkGrab}.}
We use \textsc{WalkGrab} as the main controlled ablation task because it is the cleanest test of grasping during locomotion. It uses a single-stage reward, does not use reference state initialization, and directly couples root motion, wrist placement, and finger closure. We compare CoorDex with three variants under the same environment and task-level reward. \textit{All Joint Space} removes both latent priors and predicts joint targets directly. \textit{Body Prior + Hand Joint Space} keeps the body prior but controls the hand in joint space. \textit{Monolithic Latent Residual} keeps both priors but predicts one concatenated residual vector with a single MLP. Since the action parameterizations differ, we compare task-level metrics rather than raw return.

\subsection{Task Performance Across Skills}
\label{sec:task_results}

\begin{table}[t]
    \centering
    \caption{Task performance of CoorDex. Results are evaluated over $50{,}000$ episodes collected with $10{,}000$ parallel simulation environments in Isaac Lab.}
    \vspace{5pt}
    \label{tab:task_results}
    \begin{tabular}{lcccc}
        \toprule
        Task & Success & Fall & Drop & Task-specific metric \\
        \midrule
        \textsc{WalkGrab} & 0.55 & 0.00 & 0.40 & Velocity profile, Fig.~\ref{fig:walkgrab_velocity} \\
        \textsc{OpenFridge} & 0.66 & 0.00 & -- & Door angle: 57.76 / $60^\circ$ \\
        \textsc{WalkPickTurn} & 0.89 & 0.01 & 0.10 & Min heading err.: 9.98$^\circ$ \\
        \bottomrule
    \end{tabular}
    \vspace{-18pt}
\end{table}

Table~\ref{tab:task_results} shows that the same CoorDex interface can be instantiated across all three tasks, answering \textbf{Q1}. \textsc{WalkGrab} is the most challenging because grasping must happen under continuous whole-body motion, whereas \textsc{OpenFridge} permits the robot to slow down and reposition during manipulation. \textsc{WalkPickTurn} is longer-horizon, but NoDemoRSI provides additional exploration support by resetting later stages from states discovered by the policy itself, without any expert states or action labels. Implementation details are given in Appendix~\ref{sec:impl}.

\subsection{Action-Space Variants on \textsc{WalkGrab}}
\label{sec:action_space_ablation}

We next study whether the latent-prior action interface is necessary for \textsc{WalkGrab}, answering \textbf{Q2}. This task directly tests the definition of non-stop dexterous loco-manipulation: the robot must grasp an object while it keeps moving forward. Table~\ref{tab:walkgrab_action_ablation} compares the action-space variants under the same environment and task reward.

\textbf{Metrics.}
Reach measures whether the palm enters a predefined distance threshold from the bottle. Grasp measures whether the bottle is lifted above the task threshold. Stop measures whether the humanoid slows below a velocity threshold near the bottle. Fall measures robot-fall termination.

\begin{table}[t]
    \centering
    \caption{Action-space variants on \textsc{WalkGrab} under the same PPO budget. Reach, grasp, stop, and fall are rates over evaluation episodes.}
    \label{tab:walkgrab_action_ablation}
    \begin{tabular}{lccccc}
        \toprule
        Method & Success & Reach & Grasp & Stop & Fall \\
        \midrule
        All Joint Space & 0 & 1.00 & 0.00 & 0.86 & 0.04 \\
        Body Prior + Hand Joint Space & 0 & 0.96 & 0.01 & 0.90 & 0.04 \\
        \textbf{CoorDex} & 0.55 & 1.00 & 0.55 & 0.00 & 0.00 \\
        \bottomrule
    \end{tabular}
    \vspace{-12pt}
\end{table}

\begin{figure}[t]
    \centering
    \includegraphics[
        width=\linewidth,
        trim=30pt 0pt 30pt 0pt,
        clip
    ]{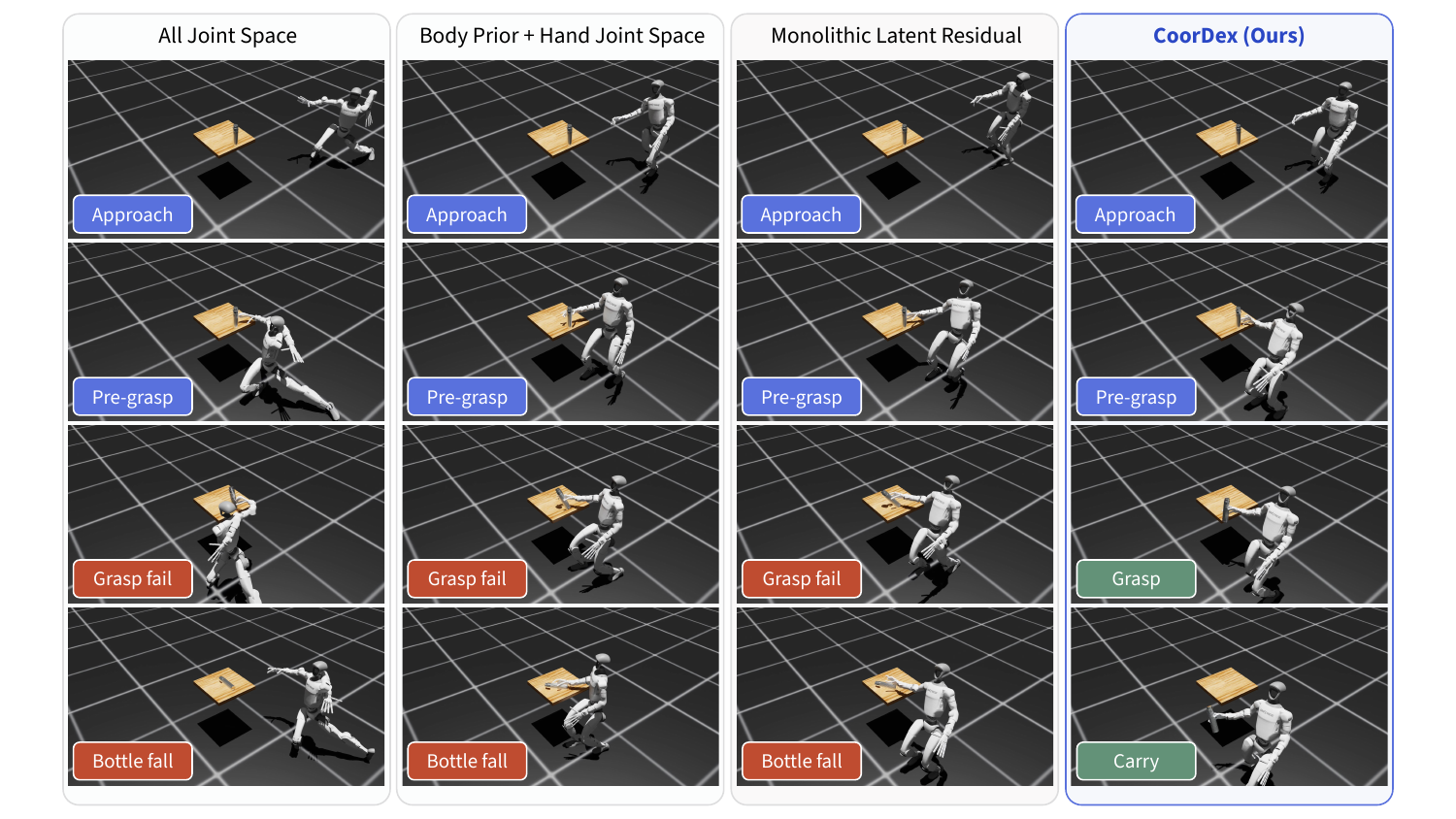}
    \vspace{-16pt}
    \caption{
        \textbf{Qualitative comparison on \textsc{WalkGrab}.}
        Each column shows sequential key frames from one rollout of the corresponding method.
        All Joint Space produces unstable whole-body motion.
        Body Prior + Hand Joint Space reaches the bottle but fails to learn a reliable grasp.
        Monolithic Latent Residual reaches the interaction region but produces less natural body motion and fails to complete the task.
        \textbf{CoorDex} completes the full sequence of approach, grasp, lift, and carry.
    }
    \vspace{-12pt}
    \label{fig:experiment_demo}
\end{figure}

The two failed variants reveal different failure modes, also visible in Fig.~\ref{fig:experiment_demo}. All Joint Space removes both learned priors and must explore the full body and hand joint space. The policy often twists the whole body into unnatural postures to avoid falling, and it never grasps the bottle.

Body Prior + Hand Joint Space isolates the hand-side difficulty. With the body prior, the humanoid can walk toward the bottle and place the wrist near the interaction region. However, the policy still has to discover high-DoF finger coordination directly in joint space. In practice, it often learns to slow down or stop near the bottle and tries to solve the problem as a stationary grasping task. This behavior shows that the body prior alone is not enough for non-stop dexterous loco-manipulation. The hand prior is also needed to make finger coordination learnable under residual RL.

For \textsc{WalkGrab}, we also analyze non-stop behavior with a velocity profile. For each rollout, we compute the body-frame forward velocity $v^{\mathrm{body}}_{x,t}$ and the relative forward position $d_t = x^{\mathrm{robot}}_t - x^{\mathrm{bottle}}_t$. We bin samples by $d_t$ with bin width $0.05\,\mathrm{m}$ over the range $[-3.0, 0.5]\,\mathrm{m}$, and plot the mean and standard deviation of $v^{\mathrm{body}}_{x,t}$ in each bin. 
\begin{wrapfigure}{r}{0.5\linewidth}
    \centering
    \vspace{-12pt}
    \includegraphics[width=\linewidth]{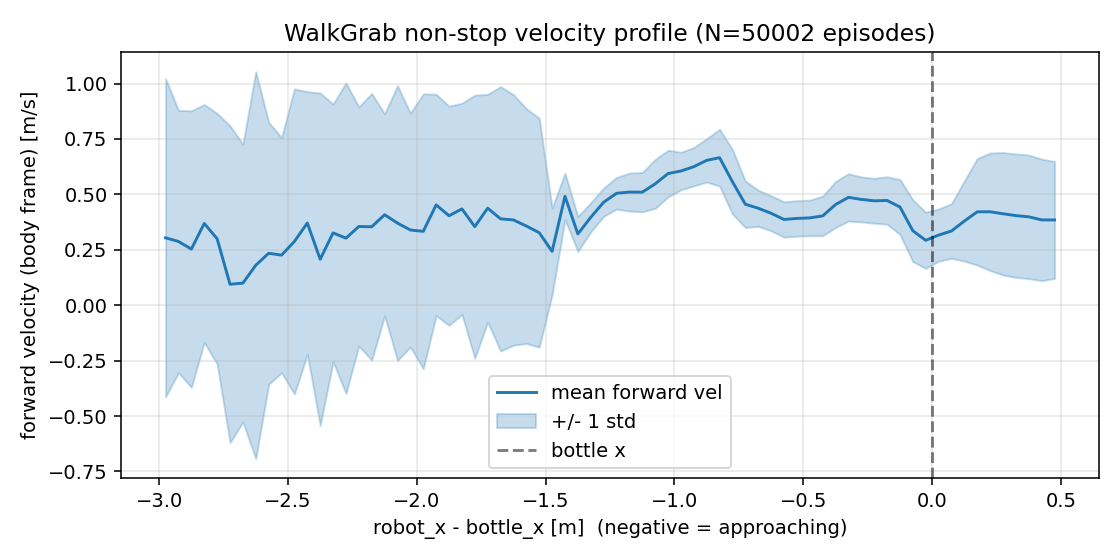}
    \vspace{-20pt}
    \caption{Non-stop locomotion on \textsc{WalkGrab}.}
    \label{fig:walkgrab_velocity}
    \vspace{-20pt}
\end{wrapfigure}
As shown in Fig.~\ref{fig:walkgrab_velocity}, CoorDex slows down slightly near the bottle but still keeps a forward velocity around $0.25\,\mathrm{m/s}$ near $d_t=0$. This indicates that the policy does not solve the task by stopping before grasping.

\subsection{Coordinated Residual Prediction}
\label{sec:coord_ablation}

We finally test whether the actor structure matters once both priors are available, answering \textbf{Q3}. Monolithic Latent Residual and CoorDex use the same frozen body and hand priors, the same latent dimensions, and the same RL environment and reward. The only difference is the policy architecture. Monolithic Latent Residual uses one MLP to predict the full latent residual and then splits it into body and hand parts. CoorDex uses a shared coordination trunk with separate body and hand residual heads.

\textbf{Metrics.}
Action rate measures the average change of decoded body joint targets between consecutive control steps. A lower action rate indicates smoother and more natural body motion.


\begin{wraptable}{r}{0.56\linewidth}
    \centering
    \vspace{-8pt}
    \caption{Coordination analysis on \textsc{WalkGrab}.}
    \label{tab:coord_ablation}
    \small
    \setlength{\tabcolsep}{4pt}
    \begin{tabular}{lccc}
        \toprule
        Method & Success & Action rate & Fall \\
        \midrule
        Monolithic Latent Residual & 0.00 & 0.40 & 0.02 \\
        \textbf{CoorDex} & \textbf{0.55} & \textbf{0.22} & 0.00 \\
        \bottomrule
    \end{tabular}
    \vspace{-8pt}
\end{wraptable}

Table~\ref{tab:coord_ablation} and Fig.~\ref{fig:experiment_demo} show that the coordinated actor matters even when both priors are present. The monolithic actor can often follow the task direction and approach the bottle, but its body motion is less natural and more jittery. This is reflected in the higher action rate in Table 4 and the unstable torso motion visible in Fig. 3. The hand failure is harder to see because the fingers are small, but the zero success rate shows that monolithic latent prediction does not produce reliable grasping under the same reward budget. CoorDex performs better because the shared trunk lets the policy reason about the same task state, while the separate heads keep body adaptation and finger adaptation from being forced through a single output pathway.

\section{Discussion and Limitations}
We present CoorDex, a modular pipeline for dexterous loco-manipulation. Rather than treating locomotion and manipulation as two separate phases, CoorDex exposes a structured latent action space for the humanoid body and the dexterous hand. Within this space, downstream RL can jointly adapt stepping, wrist placement, and finger-level contact. Across bottle grasping and carrying, fridge door opening, and cube pick-and-turn, the results indicate that dexterous humanoid skills benefit from priors that provide task-relevant motion support, and from policies that preserve the structure of body-hand coordination. This points to a practical direction for future loco-manipulation systems. Reusable motion priors should be paired with task-aware latent coordination, treating whole-body mobility and dexterous contact as coupled behaviors rather than stitched-together modules.


\textbf{Limitations and Future Work.}
Although CoorDex demonstrates continuous high-DoF dexterous loco-manipulation across multiple simulated tasks, our results are still an early step toward general humanoid dexterity. The current policies use privileged state observations, including object poses and contact signals, and do not yet address perception or visual sim-to-real transfer. The experiments also focus on a fixed G1 platform with a WUJI hand, and broader deployment will require evaluation across more objects, hands, and hardware configurations. Finally, longer-horizon tasks such as \textsc{WalkPickTurn} still rely on task-specific exploration support such as NoDemoRSI. Future work should combine the proposed latent residual interface with perception, automatic curricula, and broader task-conditioned priors.




\clearpage


\appendix

\section*{\LARGE Appendix}








\section{Motion Prior Implementation Details}
\label{app:motion_prior_implementation}

This section describes the implementation for building the body and hand priors that are later composed by the downstream residual policy. The body tracker extends the motion tracking design used by BeyondMimic~\cite{liao2025beyondmimicmotiontrackingversatile}. References are represented as joint trajectories and Cartesian body targets, the policy tracks them through normalized joint-position setpoints, and the critic receives additional privileged tracking information. The hand tracker is trained with PPO to track MANO keypoints retargeted to robot hand bodies~\cite{li2025maniptransefficientdexterousbimanual}. The distillation stage learns an encoder--decoder latent skill space together with a proprioception-conditioned prior~\cite{luo2024pulse}.

\subsection{Humanoid Body Tracking Teacher}
\label{app:body_tracking_teacher}
The robot is a Unitree G1~\cite{unitree_g1} with fixed WUJI hands~\cite{wuji_hand}. Only the 29 body joints are actuated for training efficiency. The reference bodies used by the tracking objective include the pelvis, leg links, torso, arm links, and wrist links; finger bodies are excluded from the body prior. Actions are normalized joint position setpoints with joint-specific scales derived from actuator effort and stiffness. The simulation uses decimation 4 and a physics step of $1/(60\cdot4)$ seconds, giving a 60 Hz control rate and 10 second episodes.

The no-state-estimator body teacher removes anchor translation and base linear velocity from the actor and teacher observations for real-world deployment. Per frame, the deployable proprioceptive suffix is
\[
    [\omega_{\mathrm{base}}, q, \dot{q}, a_{t-1}] \in \mathbb{R}^{90},
\]
where $q,\dot{q},a_{t-1}$ each have 29 body-joint dimensions. The actor also receives the current motion command and relative anchor orientation during teacher training, with a five-frame history. The critic is asymmetric and additionally observes privileged reference body pose and velocity errors.

Rewards are exponential tracking rewards plus minimal regularization. The main positive terms track anchor position and orientation, relative body position and orientation, body linear and angular velocity, and palm keypoints. Penalties discourage action rate jitter, joint limit violations, and undesired contacts. Episodes terminate on timeout, excessive anchor height or orientation error, or excessive end-effector height error at the ankles and wrist yaw links. Domain randomization includes friction, joint default offsets, torso center-of-mass perturbation, and random base pushes.

The teacher is trained with PPO using 24 rollout steps per environment, five learning epochs, four minibatches, $\gamma=0.99$, $\lambda=0.95$, adaptive learning rate with target KL 0.01, learning rate $10^{-3}$, entropy coefficient 0.005, and gradient clipping at 1.0. Actor and critic MLPs use ELU activations and hidden dimensions $[1024,512,256]$.

\subsection{Dexterous Hand Tracking Teacher}
\label{app:hand_tracking_teacher}
Our hand motion loader reads ManipTrans-style annotations~\cite{li2025maniptransefficientdexterousbimanual}, evaluates SMPL-X to obtain MANO wrist and finger keypoints, converts from the MuJoCo frame to the Isaac Lab frame, and downsamples 120 Hz annotations to 60 Hz.

For the current WUJI hand prior, the tracked hand bodies are the palm and the five finger links and tips. The environment writes the reference wrist root pose and velocity into simulation both at command resampling and during every command update, so the wrist is kinematically driven by the dataset. The policy therefore controls only the twenty active finger joints for WUJI hand. This fingers-only action space avoids spending latent capacity on high-variance wrist motion and matches the downstream residual interface, where the humanoid body prior determines the wrist motion.

The policy observation concatenates a target command, hand proprioception, and the previous action. The target command contains flattened MANO keypoint positions and velocities, plus the reference wrist pose expressed relative to the simulated wrist. Proprioception contains wrist linear and angular velocity in the wrist frame, and active joint position and velocity relative to default joint states. The critic receives the same target and proprioception plus privileged per-body keypoint position and velocity deltas.

The remaining rewards track MANO keypoints with fingertip-heavy weights, track keypoint velocities, penalize joint power through an exponential power reward, and penalize action rate. Tracking failure terminates an episode if keypoint errors exceed group-specific thresholds after a short warmup. An additional unstable state termination catches unrealistic wrist, body, or joint velocities. PPO hyperparameters match the body teacher: 24 rollout steps, five epochs, four minibatches, ELU MLPs with $[1024,512,256]$ hidden units, learning rate $10^{-3}$, adaptive KL target 0.01, and 4096 environments.

\subsection{VAE Distillation}
\label{app:vae_distillation}
The student contains three trainable networks:
\[
    q_\phi(z_t\mid s^{\mathrm{full}}_t), \qquad
    p_\psi(z_t\mid s^{\mathrm{prop}}_t), \qquad
    D_\theta(s^{\mathrm{prop}}_t,z_t).
\]
The encoder predicts the posterior diagonal Gaussian from the full student observation. The prior predicts a diagonal Gaussian from the deployable proprioceptive slice. And the decoder maps proprioception plus a latent sample to the teacher action. During training, $z_t$ is sampled from the encoder with the reparameterization trick. At inference and downstream residual RL time, the default latent is the prior mean.

The distillation loss is
\[
    \mathcal{L}
    =
    \lambda_a \|\hat{a}_t-a^T_t\|_2^2
    + \lambda_s \|\mu^q_t-\mu^q_{t-1}\|_2^2
    + \lambda_{\mathrm{KL}} D_{\mathrm{KL}}
    \left(q_\phi(z_t\mid s^{\mathrm{full}}_t)\,\|\,p_\psi(z_t\mid s^{\mathrm{prop}}_t)\right),
\]
where the smoothness term is evaluated only across valid consecutive samples in the same environment. This mirrors the PULSE~\cite{luo2024pulse} motivation. The decoder learns the teacher's motor actions. The prior makes the latent usable from proprioception alone. And the temporal regularizer keeps nearby states from mapping to discontinuous latent codes.

For the body prior, the latent dimension is 16.  The proprioceptive prior input is the five-frame suffix
\[
    5 \times [\omega_{\mathrm{base}}(3), q(29), \dot{q}(29), a_{t-1}(29)]
    = 450
\]
dimensions. Encoder, decoder, and prior networks are MLPs with $[512,256,128]$. The body distillation optimizer uses learning rate $2\cdot10^{-4}$, action coefficient 1.0, smoothness coefficient 0.005, and a KL coefficient annealed from $10^{-3}$ to $10^{-4}$ between 15000 and 20000 updates.

For the active WUJI hand prior, the latent dimension is 12. The prior input is 66 dimensions:
\[
    [v_{\mathrm{wrist}}(3), \omega_{\mathrm{wrist}}(3), q_{\mathrm{rel}}(20),
    \dot{q}_{\mathrm{rel}}(20), a_{t-1}(20)].
\]
The decoder outputs twenty fingers-only actions. The hand distillation optimizer uses learning rate $5\cdot10^{-4}$, action coefficient 1.0, smoothness coefficient 0.005, and a KL coefficient annealed from $10^{-2}$ to $10^{-3}$ between 15000 and 25000 updates. Both distillation runners use 24 rollout steps, five learning epochs, gradient accumulation length 16, gradient clipping at 1.0, empirical observation normalization, and up to 30000 iterations.

\subsection{Downstream Residual Composition}
\label{app:downstream_residual_composition}

Downstream loco-manipulation tasks use the frozen priors. The action term loads only the distilled prior and decoder weights, plus the observation normalizer. At each control step it constructs a body-prior observation from the current G1 proprioceptive history and a hand-prior observation from wrist-frame hand velocity, finger joint state, and the previous decoded hand action.

The downstream PPO actor outputs a residual latent
\[
    \Delta z_t = [\Delta z^b_t,\Delta z^h_t].
\]
For each prior, the action term computes the prior mean and decodes the shifted latent:
\[
    a^b_t = D_b(o^b_t,\mu^b_t+\Delta z^b_t), \qquad
    a^h_t = D_h(o^h_t,\mu^h_t+\Delta z^h_t).
\]
The decoded body actions are inserted into the 29 body-joint slots. The decoded hand actions are converted through the configured joint-action scale and offset into finger targets. An EMA with coefficient 0.4 is applied to the hand targets. The combined joint position action is then sent to the underlying Isaac Lab joint position action term.

The body prior is a 16-dimensional no-base-linear-velocity G1 prior and the hand prior is the 12-dimensional kinematic-wrist WUJI hand prior, yielding a 28-dimensional residual action space.

\section{Training Details}
\label{sec:impl}

We instantiate three downstream loco-manipulation tasks on the same Unitree~G1
humanoid equipped with a WUJI five-finger right hand: \textbf{\textsc{WalkGrab}}, \textbf{\textsc{OpenFridge}}, and \textbf{\textsc{WalkPickTurn}}. All
three share the same control stack. The policy runs at $60$~Hz (physics step
$1/240$~s, decimation $4$) over $4096$ parallel environments, and emits a $28$-dim
latent residual ($16$ body $+\,12$ hand) on top of two frozen priors mentioned above. \textsc{WalkGrab} and \textsc{OpenFridge} are single-stage reformulations whose phase structure is realized through
predicate gating inside the reward terms, whereas \textsc{WalkPickTurn} keeps an explicit
three stage schedule with a stage weighted reward.

\subsection{Observation Details}
\label{sec:impl-obs}

For every task the actor and critic observe the same terms. The only difference is
that observation noise is enabled for the actor and disabled for the
critic. Each observation concatenates the proprioceptive prior stack
(\texttt{humanoid\_prior}, a $5$-step history of the $90$-dim body proprio frame
$[\,v^{\text{base}}_{\text{ang}}(3),\,q(29),\,\dot q(29),\,a_{t-1}(29)\,]$;
\texttt{right\_hand\_prior}, the $66$-dim hand proprio frame; and the last emitted
latent), the frozen prior latent means, contact forces, and task-specific object
poses. Object poses are expressed as a $3$-D position plus the first two columns of
the rotation matrix ($6$-D), i.e. $9$ dims each. Table~\ref{tab:obs-all} lists the per-task layouts.

\begin{table*}[t]
\centering
\caption{Observation terms for the three downstream tasks. Policy and critic share the same layout.}
\label{tab:obs-all}
\scriptsize
\setlength{\tabcolsep}{3pt}

\begin{minipage}[t]{0.33\textwidth}
\centering
\caption*{\textsc{WalkGrab}}
\begin{tabular}{lc}
\toprule
\textbf{State term} & \textbf{Dim.} \\
\midrule
Stage one-hot                  & 1   \\
Projected gravity              & 3   \\
Body prior proprio history     & 450 \\
Right-hand prior proprio       & 66  \\
Last latent action             & 28  \\
Fingertip forces on bottle     & 15  \\
Bottle pose, root              & 9   \\
Source-table pose, root        & 9   \\
Bottle pose, hand              & 9   \\
Body prior latent mean         & 16  \\
Hand prior latent mean         & 12  \\
\midrule
Total                          & 618 \\
\bottomrule
\end{tabular}
\end{minipage}
\hfill
\begin{minipage}[t]{0.33\textwidth}
\centering
\caption*{\textsc{WalkPickTurn}}
\begin{tabular}{lc}
\toprule
\textbf{State term} & \textbf{Dim.} \\
\midrule
Stage one-hot                  & 3   \\
Projected gravity              & 3   \\
Body prior proprio history     & 450 \\
Right-hand prior proprio       & 66  \\
Last latent action             & 28  \\
Fingertip forces on object     & 15  \\
Object pose, root              & 9   \\
Source-table pose, root        & 9   \\
Object pose, hand              & 9   \\
Body prior latent mean         & 16  \\
Hand prior latent mean         & 12  \\
\midrule
Total                          & 620 \\
\bottomrule
\end{tabular}
\end{minipage}
\hfill
\begin{minipage}[t]{0.33\textwidth}
\centering
\caption*{\textsc{OpenFridge}}
\begin{tabular}{lc}
\toprule
\textbf{State term} & \textbf{Dim.} \\
\midrule
Stage one-hot                  & 1   \\
Projected gravity              & 3   \\
Body prior proprio history     & 450 \\
Right-hand prior proprio       & 66  \\
Last latent action             & 28  \\
Fingertip forces on handle     & 15  \\
Fridge pose, root              & 9   \\
Handle-point pose, root        & 9   \\
Door pose, root                & 9   \\
Handle-point pose, hand        & 9   \\
Door joint state               & 2   \\
Body prior latent mean         & 16  \\
Hand prior latent mean         & 12  \\
\midrule
Total                          & 629 \\
\bottomrule
\end{tabular}
\end{minipage}

\end{table*}

\subsection{Reward Details}
\label{sec:impl-rew}

\paragraph{\textsc{WalkGrab} and \textsc{OpenFridge} (single-stage).}
Both tasks use a single ``continuous'' stage, so the per-step reward is a flat
weighted sum of shaping terms,
\begin{equation}
r_t \;=\; \sum_{k} w_k\, r^{(k)}_t ,
\end{equation}
where phase structure is induced by predicate gating \emph{inside} the terms:
a grasp predicate $\mathds{1}[\text{grasped}]$ turns approach terms off and
manipulation terms on, and a contact predicate gates the door-opening / lifting
rewards so the robot cannot exploit them without touching the object.
Tables~\ref{tab:rew-walkgrab} and \ref{tab:rew-fridge} list the active terms.

\paragraph{\textsc{WalkPickTurn} (multi-stage).}
A \textsc{WalkPickTurn} episode is decomposed into stages. Table~\ref{tab:rew-walkpickturn} instantiates $r^{(s)}$ with the stage-dependent
shaping terms. More details can be found in~\ref{sec:nodemo-rsi-stages}
\begin{equation}
r_t \;=\; \sum_{i=0}^{2} w_i\, \mathds{1}[\,s_t = i\,]\, r^{(i)}_t , \qquad w_i > 0,
\end{equation}

\begin{table}[t]
\centering
\caption{\textsc{WalkGrab} reward components, expressions, and weights. $\overline{d}$ denotes
a mean distance over the listed bodies; \emph{grasped} requires mean fingertip--bottle distance $<0.10$ m and at least two fingertip contact sensors above 1 N, including thumb contact.}
\label{tab:rew-walkgrab}
\begin{tabular}{p{0.30\linewidth} p{0.46\linewidth} r}
\toprule
\textbf{Term} & \textbf{Expression} & \textbf{Weight} \\
\midrule
\multicolumn{3}{l}{\emph{Task / object-centric rewards}}\\
Hand--bottle distance   & $\tfrac12\big[\exp(-8.5\,\overline{d}_{\text{palm+tips}})$ & $50.0$ \\
Grasp by finger dir     & $\theta_{\text{thumb,others}}/\pi$ & $40.0$ \\
Fingertip contact       & $\text{mean}_i\,\mathds{1}[\lVert F_i\rVert > 1]$ & $15.0$ \\
Grasp force             & $\mathrm{clip}\!\big(\text{mean}_i \textstyle\sum\lVert F_i\rVert,\,0,2\big)$ & $15.0$ \\
Palm facing object      & $\exp\!\big(-((\cos\angle_{xy}-0.47)/0.25)^2\big)$ & $15.0$ \\
Thumb opposition        & $\sum_i w_i\exp(-|q_i-q^\ast_i|)$ (4 thumb joints) & $20.0$ \\
Others open (penalty)   & $[1-\exp(-3\,\overline{\text{dev}})]+5\,\mathrm{clip}(\Delta\text{err},0,0.02)$ & $-5.0$ \\
Bottle lift height      & $(\min(\ell,0.05)/0.05)\exp(-2\,v_z^2)$, grasp-gated & $100.0$ \\
Hold bottle             & $\mathds{1}[\text{grasped}]$ & $100.0$ \\
Sustained grasp         & $\mathrm{clip}(\text{secs}(\text{grasped}\wedge\ell>0.01)/1.5,0,1)$ & $100.0$ \\
Forward progress        & $\mathrm{clip}(v_x^{\text{bottle}}/1,0,1)\,\mathds{1}[\text{grasped}\wedge\ell>0.01]$ & $100.0$ \\
\midrule
\multicolumn{3}{l}{\emph{Locomotion / posture shaping}}\\
Forward target          & $\exp\!\big(-[(x-2)^2+y^2]\big)$ & $25.0$ \\
Heading forward         & $\exp(-(\Delta\psi)^2/0.3^2)$ & $10.0$ \\
Upright orientation     & $\exp(-\lVert g_{xy}\rVert^2/0.2^2)$ & $5.0$ \\
$y$ below zero (penalty)& $\mathrm{clip}(-y,0,0.5)$ & $-10.0$ \\
\midrule
\multicolumn{3}{l}{\emph{Termination / generic penalties}}\\
Success bonus           & $\mathds{1}[\text{success}]$ & $500.0$ \\
Action rate             & $\lVert a_t-a_{t-1}\rVert^2$ & $-0.01$ \\
Undesired table contact & $\textstyle\sum \mathds{1}[\text{contact}>5\,\text{N}]$ & $-20.0$ \\
Passed without grasp    & $\mathds{1}[\text{passed-without-grasp}]$ & $-20.0$ \\
Time-out                & $\mathds{1}[\text{time-out}]$ & $-20.0$ \\
Robot fall              & $\mathds{1}[\text{fall}]$ & $-10.0$ \\
Bottle dropped          & $\mathds{1}[\text{dropped}]$ (scaled $\times5$ once lifted) & $-1.0$ \\
\bottomrule
\end{tabular}
\end{table}

\begin{table}[t]
\centering
\caption{\textsc{OpenFridge} reward components, expressions, and weights. $\Delta\theta$ is the
door open angle, $c\in\{0,1\}$ the fingertip-contact predicate; the success angle is
$60^\circ$ and the held-open angle $35^\circ$.}
\label{tab:rew-fridge}
\begin{tabular}{p{0.30\linewidth} p{0.46\linewidth} r}
\toprule
\textbf{Term} & \textbf{Expression} & \textbf{Weight} \\
\midrule
\multicolumn{3}{l}{\emph{Task / object-centric rewards}}\\
Hand--handle distance   & $\exp(-10\,\overline{d})$ over \{5 tips $+$ palm\} & $30.0$ \\
Door open amount        & $\mathrm{clip}(\Delta\theta/60^\circ,0,1)\cdot c$ & $30.0$ \\
Door open progress      & $\mathrm{clip}(\max(\dot\theta,0)/0.02,0,1)\cdot c$ & $20.0$ \\
Door held open          & $\mathrm{clip}(\text{secs}(\Delta\theta\ge35^\circ\wedge c)/1,0,1)$ & $100.0$ \\
Door close regression   & $\mathrm{clip}(\max(-\dot\theta,0)/0.02,0,1)$ & $-20.0$ \\
Backward progress       & $\mathrm{clip}(v_{\text{back}}/0.10,0,1)\cdot c$ & $20.0$ \\
Fingertip contact       & $\text{mean}_i\mathds{1}[f_i>1]+0.25(\tfrac{\max(n_c-2,0)}{3})^2$ & $15.0$ \\
Thumb contact           & $\mathds{1}[f_{\text{thumb}}>1]$ & $10.0$ \\
\midrule
\multicolumn{3}{l}{\emph{Termination / generic penalties}}\\
Success bonus           & $\mathds{1}[\text{success}]$ & $1000.0$ \\
Action rate             & $\mathrm{clip}(\lVert a_t-a_{t-1}\rVert^2,\le100)$ & $-0.01$ \\
Joint action rate       & $\mathrm{clip}(\sum(u_t-u_{t-1})^2,\le50)$ & $-0.01$ \\
DoF velocity            & $\mathrm{clip}(\sum_j\dot q_j^2,\le200)$ & $-0.01$ \\
Feet slip               & $\mathrm{clip}(\sum_{\text{feet}}\mathds{1}[F>5]\lVert v_{xy}\rVert,\le2)$ & $-5.0$ \\
Large linear $v_x$      & $\min(\max(|v_x|-0.8,0),2)$ & $-8.0$ \\
Large linear $v_y$      & $\min(\max(|v_y|-0.35,0),2)$ & $-10.0$ \\
Large angular $\omega$  & $\min(\max(|\omega_z|-0.6,0),5)$ & $-10.0$ \\
Hand far from handle     & $\mathds{1}[\text{hand far}]$ & $-20.0$ \\
Time-out                & $\mathds{1}[\text{time-out}]$ & $-20.0$ \\
Robot fall              & $\mathds{1}[\text{fall}]$ & $-10.0$ \\
\bottomrule
\end{tabular}
\end{table}

\begin{table*}[t]
\centering
\caption{\textsc{WalkPickTurn} reward components, expressions, weights, and the stage(s)
($0$--$2$) where each term is applied. Stages: (0) approach \& hand-prep, (1) grasp
\& lift, (2) turn.}
\label{tab:rew-walkpickturn}
\begin{tabular}{p{0.26\linewidth} p{0.44\linewidth} r c}
\toprule
\textbf{Term} & \textbf{Expression} & \textbf{Weight} & \textbf{Stage(s)} \\
\midrule
\multicolumn{4}{l}{\emph{Termination / generic penalties}}\\
Success bonus            & $\mathds{1}[\text{success}]$ & $1000.0$ & --- \\
Robot fall               & $\mathds{1}[\text{fall}]$ & $-10.0$ & --- \\
Action rate              & $\lVert a_t-a_{t-1}\rVert^2$ & $-0.01$ & 0--2 \\
Joint action rate        & $\sum(u_t-u_{t-1})^2$ & $-0.01$ & 0--2 \\
Left-arm DoF velocity    & $\sum\dot q_{\text{l-arm}}^2$ & $-0.01$ & 0--2 \\
DoF velocity             & $\sum_j\dot q_j^2$ & $-0.01$ & 0--2 \\
Large linear $v_x$       & $\mathrm{clip}(|v_x|-0.5,0,5)$ & $-10.0$ & 0--2 \\
Large linear $v_y$       & $\mathrm{clip}(|v_y|-0.5,0,5)$ & $-10.0$ & 0--2 \\
Large angular $\omega$   & $\mathrm{clip}(|\omega_z|-0.5,0,5)$ & $-5.0$ & 0--2 \\
Feet slip                & $\sum_{\text{feet}}\mathds{1}[\text{contact}]\lVert v_{xy}\rVert$ & $-5.0$ & 0--2 \\
Undesired table contact  & $\textstyle\sum\mathds{1}[\text{contact}>5\,\text{N}]$ & $-10.0$ & 0--2 \\
Upright orientation      & $\exp(-\lVert g_{xy}\rVert^2/0.2^2)$ & $5.0$ & 0--2 \\
\midrule
\multicolumn{4}{l}{\emph{Heading / command shaping}}\\
Stage-progress bonus     & $\mathds{1}[s_t>s_{t-1}]$ & $500.0$ & 0--2 \\
Turn alignment           & $-|\Delta\psi_{\text{region}}|$ & $15.0$ & 2 \\
Heading to object        & $(\Delta\psi_{\text{obj}}/\pi)^2$ & $-10.0$ & 0 \\
\midrule
\multicolumn{4}{l}{\emph{Task / object-centric rewards}}\\
Robot--object distance   & $\exp(-4(d-0.5)^2)$ & $15.0$ & 0--2 \\
Right-hand target pose   & $\exp(-25\lVert\bar p_{\text{hand}}-p^\ast\rVert^2)$ & $30.0$ & 0 \\
Hand prep-region penalty & $1-\exp(-(60\,e_{xy}^2+20\,e_z^2))$ & $-10.0$ & 0 \\
Hand flat (stage 0)      & $1-\exp(-(\alpha/0.8)^2)$ & $-5.0$ & 0 \\
Keep hand open           & $\exp(-6\,\overline{\lVert q-q_{\text{open}}\rVert})+5\,\text{prog}$ & $20.0$ & 0 \\
Hand--object distance    & $\exp(-10\,\overline{d})$ over \{5 tips $+$ palm\} & $20.0$ & 1--2 \\
Fingertip contact        & contact ratio $+$ count bonus, lift-gated & $10.0$ & 1--2 \\
Grasp force              & $\tanh\!\big(\textstyle\sum\lVert F\rVert_{\text{clip}}/2\big)$, lift-gated & $10.0$ & 1--2 \\
Object lift height       & $\min(\ell,0.12)/0.12$ & $50.0$ & 1--2 \\
Hold object              & $\mathds{1}[\text{grasped}],\mathds{1}[\ell>\ell_0]$ & $20.0$ & 1--2 \\
Object not lifted (pen.) & $1-\mathrm{clip}(\ell/0.1,0,1)$ & $-10.0$ & 1--2 \\
\bottomrule
\end{tabular}
\end{table*}

\subsection{Hyperparameter Details}
\label{sec:impl-hp}

All three tasks are trained with PPO under the same actor--critic architecture and
optimizer. Only the task-dependent settings (episode length, number of stages, and
initial action noise) differ. The actor is a coordinated residual network with a shared coordination trunk that feeds body and hand heads that each output a latent residual
on the corresponding frozen prior. The shared PPO settings are listed in
Table~\ref{tab:hp-shared} and the per-task differences in Table~\ref{tab:hp-task}.

\begin{table}[t]
\centering
\caption{Shared training hyperparameters (identical across all three tasks).}
\label{tab:hp-shared}
\begin{tabular}{lc}
\toprule
\textbf{Hyperparameter} & \textbf{Value} \\
\midrule
Parallel environments               & 4096 \\
Steps per environment per rollout    & 24 \\
Batch size (env $\times$ steps)      & 98304 \\
Mini-batches                         & 4 \\
Learning epochs per update           & 5 \\
Discount $\gamma$                    & 0.99 \\
GAE $\lambda$                        & 0.95 \\
Clip parameter                       & 0.2 \\
Entropy coefficient                  & 0.005 \\
Value loss coefficient               & 1.0 \\
Learning rate (adaptive, KL target)  & $1\times10^{-3}$ \\
Desired KL                           & 0.01 \\
Max gradient norm                    & 1.0 \\
Empirical observation normalization  & enabled \\
Actor / critic hidden dims           & [1024, 512, 256] \\
Activation                           & ELU \\
Coord. trunk hidden dims             & [512, 256] \\
Body / hand head hidden dims         & [256, 128] \\
Body / hand residual scale           & 1.0 \\
Body / hand prior latent dim         & 16 / 12 \\
\bottomrule
\end{tabular}
\end{table}

\begin{table}[t]
\centering
\caption{Task-dependent settings.}
\label{tab:hp-task}
\begin{tabular}{lccc}
\toprule
\textbf{Setting} & \textbf{\textsc{WalkGrab}} & \textbf{\textsc{OpenFridge}} & \textbf{\textsc{WalkPickTurn}} \\
\midrule
Episode length (s)   & 15.0 & 7.0  & 10.0 \\
Control steps        & 900  & 420  & 600  \\
Number of stages     & 1    & 1    & 3    \\
Action dim (latent)  & 28   & 28   & 28   \\
Observation dim      & 618  & 629  & 620  \\
Init.\ action noise $\sigma$ & 0.22 & 0.22 & 0.22 \\
\bottomrule
\end{tabular}
\end{table}

\subsection{Task-specific success conditions and auxiliary metrics.}
\label{tab:task_success}
\begin{table}[H]
\centering
\begin{tabular}{p{0.16\linewidth}p{0.43\linewidth}p{0.33\linewidth}}
\toprule
Task & Success condition & Auxiliary metrics \\
\midrule
\textsc{WalkGrab} &
Root forward displacement $>2.0$ m, bottle lift $>0.02$ m, and grasp is true.  &
Root $x$ progress, fingertip--bottle distance, contact/grasp rate, bottle lift, carry-forward distance, bottle drop, pass-without-grasp, robot fall. \\
\midrule
\textsc{OpenFridge} &
Door angle $\ge 60^\circ$, held-open counter $\ge 30$ control steps. The held-open counter increments only while door angle $\ge 35^\circ$ and at least one fingertip contact sensor is above 1 N. &
Final/max door angle, held-open seconds, hand--handle distance, fingertip contact, backward distance, hand-far termination, robot/joint/fridge blow-up. \\
\midrule
\textsc{WalkPickTurn} &
Grasp the cube with minimum fingertip--cube distance $<0.08$ m and at least one fingertip contact $>1$ N, lifting the cube by $>0.10$ m, and completing the turn with heading error $<0.4$ rad. &
Max stage reached, stage-transition rates, fingertip--cube distance, fingertip contact, cube lift, heading error, cube drop. \\
\bottomrule
\end{tabular}
\end{table}

\subsection{Demonstration-Free Reference State Initialization}
\label{sec:nodemo-rsi}

Long-horizon loco-manipulation suffers from a sparse exploration bottleneck. A
policy that starts every episode from a default standing pose almost never
discovers the later phases of the task by chance, because reaching them requires
first solving every earlier phase. A common remedy is reference state
initialization (RSI) in VIRAL~\cite{viral2025}, which resets a fraction of the environments to states drawn
from a demonstration so that later phases receive gradient signal from the outset.
\textsc{WalkPickTurn} has no such demonstrations available. We therefore use a
\emph{demonstration-free} variant (NoDemoRSI) that bootstraps its own reset
distribution from states the policy visits during training, rather than from an
external dataset.

\paragraph{Stage Decomposition for \textsc{WalkPickTurn}}
\label{sec:nodemo-rsi-stages}

We decompose a \textsc{WalkPickTurn} episode into three stages, indexed $0$ to $2$, and treat
\emph{completing stage~2 as task success}. The policy controls the G1 humanoid at
$60$~Hz over episodes of $10$~s ($600$ control steps), and a per-environment stage
index advances monotonically as the corresponding sub-goal is met.

\begin{description}
  \item[Stage 0 (approach and reach).] The robot walks toward the object and brings
  its open right hand above it. The stage completes when the mean height of the
  palm and fingertips is within $0.3$~m above the object, the mean hand-to-object
  horizontal distance is below $0.1$~m, the right hand is open, both feet are in a
  stable stance (foot stagger below $0.4$~m), and no table collision is detected.
  \item[Stage 1 (grasp and lift).] The robot closes the hand on the object and lifts
  it. The stage completes when the object is grasped, defined as a minimum
  fingertip-to-object distance below $0.08$~m together with a positive fingertip
  contact, and the object is raised more than $0.1$~m above its initial height.
  \item[Stage 2 (turn).] While holding the object, the robot rotates its base to
  face the target heading. The stage completes when the heading error to the target
  falls below $0.4$~rad.
\end{description}

The transition criteria are one-way. An environment that satisfies the gate for
stage $k\!\to\!k\!+\!1$ has its stage index incremented and never decreases it
within an episode. This monotone structure is what lets NoDemoRSI treat the entry
state of each stage as a reusable initialization point.

\paragraph{Self-Populating Snapshot Buffer}
\label{sec:nodemo-rsi-buffer}

NoDemoRSI maintains one fixed-capacity ring buffer per non-initial stage. Each entry
is a \emph{snapshot} of the full simulator state needed to resume an episode from a
stage boundary. It includes the robot root state (position, orientation quaternion, and linear
and angular velocities), the positions and velocities of all actuated joints, the
object root state, the stage index, and the object's initial height. Each buffer
holds up to $512$ snapshots. Once full, new snapshots overwrite the oldest in
first-in-first-out order, so the buffer tracks the policy's current behaviour rather
than its early, low-quality behaviour.

Snapshots are collected on the fly. When an environment first transitions into a
stage $k \ge 1$, a deferred capture is queued for that environment. On the following
control step, provided the environment is still in stage $k$, the current state is
captured and inserted into buffer $k$, but only if it passes a set of quality
filters that reject states unsuitable as reset points. A snapshot is stored only
when the robot has not fallen (root height at least $0.5$~m), the environment has
spent a minimum dwell time in the stage (12 steps for stage~1, 6 steps for stages~2), and the motion is settled rather than mid-transient: root linear velocity
below $0.4$~m/s, root angular velocity below $0.75$~rad/s, joint velocity norm below
$10$~rad/s, and object linear and angular velocities below $1.0$ in their respective
units.

\paragraph{Reset Sampling and Progressive Stage Unlocking}
\label{sec:nodemo-rsi-sampling}

At every reset, NoDemoRSI assigns each resetting environment a target stage. An
environment assigned to stage~0 is reset to the default standing pose, while an
environment assigned to a later stage is reset by sampling a snapshot from that
stage's buffer and writing it back into the simulator, so training resumes from a
state the policy itself previously reached. If a later stage is selected but its
buffer is empty, the reset falls back to stage~0.

The per-stage sampling probabilities are computed adaptively. Each stage receives a
weight proportional to its difficulty times its availability. Difficulty is
$\max(1 - \mathrm{SR}_k,\, \epsilon)^{p}$ with floor $\epsilon = 0.1$ and power
$p = 1$, where $\mathrm{SR}_k$ is a smoothed success rate for resets at stage~$k$. A
reset counts as a success if the resulting episode advances to the next stage (or,
for the final stage, reaches the task success heading condition). The success rate is tracked
as an exponential moving average with decay $0.98$ and a Beta-style prior of weight
$4$ centred at $0.5$, which keeps the estimate stable when a stage has few recent
attempts. Availability is a log-compressed function of the buffer occupancy, so a
stage with more stored snapshots is sampled more readily without dominating the
difficulty term. When all tracked stages reach a success rate of $0.80$, the sampler
enters a consolidation mode that spends $85\%$ of resets on stage~0 to rehearse the
full chain end to end, and exits when any stage falls back below $0.75$. In the
non-adaptive fallback, stage~0 instead receives a fixed share of $0.3$ and the
remaining mass is split across later stages by the weights $(2, 1)$ for
stages~1 through~2.

Two mechanisms govern when the sampler may reset into a later stage, which is
essential to avoid initializing a stage before the policy can make use of it. First,
a stage must have accumulated at least $128$ snapshots in its buffer before it is
eligible. Second, a progressive warmup unlocks stages in order of their success. A
stage becomes available only after the preceding stage is itself available, the
preceding stage's success rate exceeds $0.70$, and the stage's buffer satisfies the
minimum-sample requirement. Immediately after a stage unlocks, its sampling
probability is capped and ramped from an initial share of $0.05$ up to its full
adaptive value over the first $1000$ cumulative attempts at that stage. The cap
prevents a newly unlocked stage from abruptly absorbing the reset budget and
starving the stages the policy has not yet consolidated. Together these rules induce
a curriculum that begins almost entirely from the default pose, admits stage~1
resets once the policy reliably reaches and grasps the object, and only later admits
turn-stage resets, all without any external demonstration.

\begin{table}[t]
\centering
\caption{NoDemoRSI configuration for the \textsc{WalkPickTurn} run. Per-stage entries are
listed for stages $0$--$2$; stage~0 has no buffer and is always reset to the default
pose.}
\label{tab:nodemo-rsi-cfg}
\begin{tabular}{lc}
\toprule
\textbf{Setting} & \textbf{Value} \\
\midrule
Buffer capacity per stage              & 512 \\
Min.\ snapshots to unlock a stage      & 128 (stages 1--2) \\
Min.\ dwell steps before storing       & (--, 12, 6) \\
Snapshot root-height gate              & $\ge 0.5$~m \\
Max.\ root linear / angular velocity   & $0.4$~m/s\,/\,$0.75$~rad/s \\
Max.\ joint velocity norm              & $10$~rad/s \\
Max.\ object linear / angular velocity & $1.0$\,/\,$1.0$ \\
\midrule
Adaptive stage sampling                & enabled \\
Success-rate EMA decay                 & 0.98 \\
Success-rate prior (weight @ 0.5)      & 4.0 \\
Difficulty floor / power               & 0.1\,/\,1.0 \\
Consolidation enter / exit             & 0.80\,/\,0.75 \\
Consolidation stage-0 share            & 0.85 \\
Non-adaptive stage-0 share             & 0.30 \\
Non-adaptive stage weights             & (--, 2, 1) \\
\midrule
Warmup unlock mode                     & previous-stage success \\
Previous-stage success threshold       & 0.70 \\
Ramp initial share / attempts          & 0.05\,/\,1000 \\
\bottomrule
\end{tabular}
\end{table}

\section{Real-World Demos}
\begin{figure}[t]
    \centering
    \includegraphics[width=\linewidth]{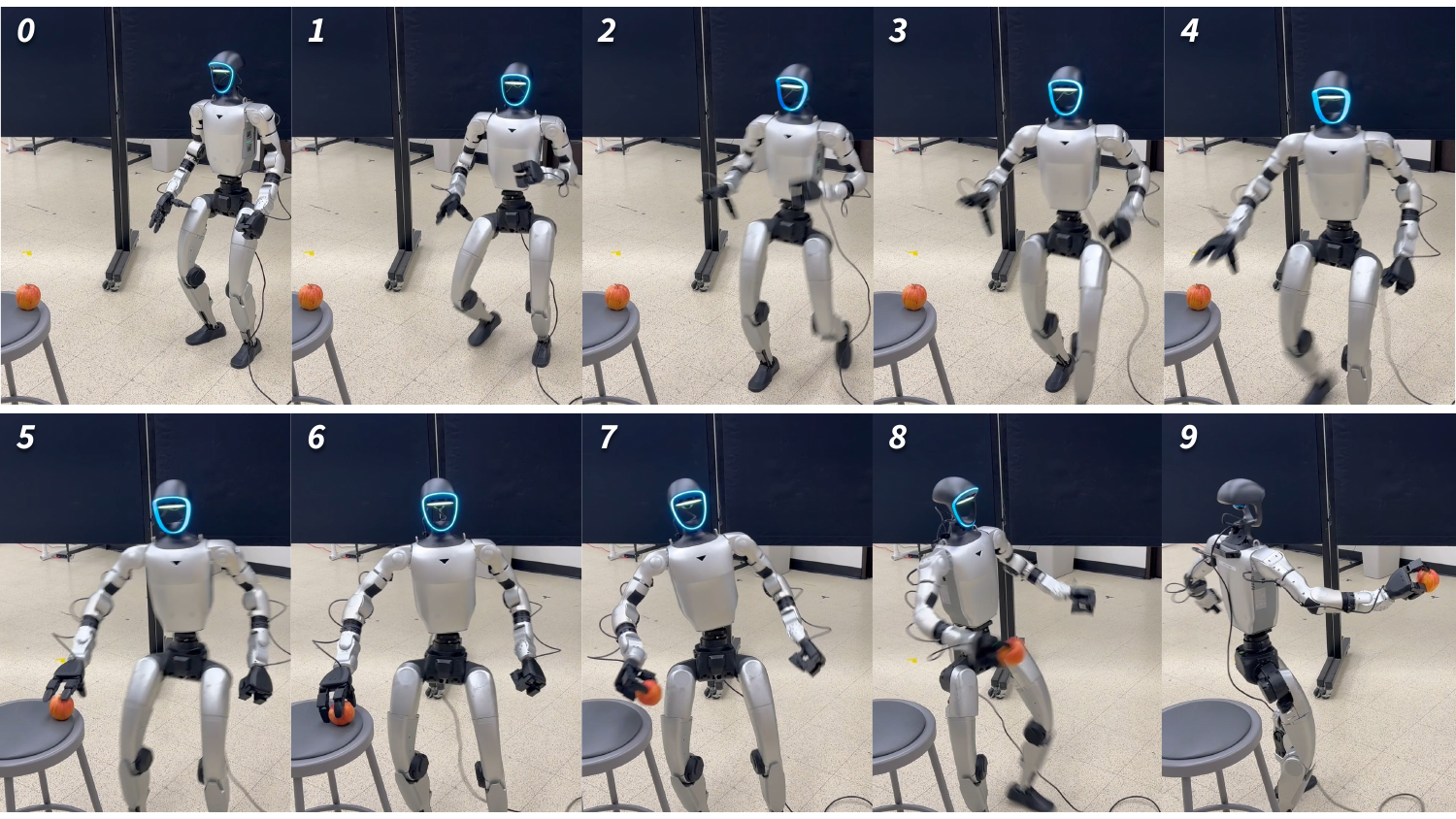}
    \caption{
    \textsc{WalkPickTurn} real-world demo.
    }
    \label{fig:walkpickturn_supply}
\end{figure}

\begin{figure}[t]
    \centering
    \includegraphics[width=\linewidth]{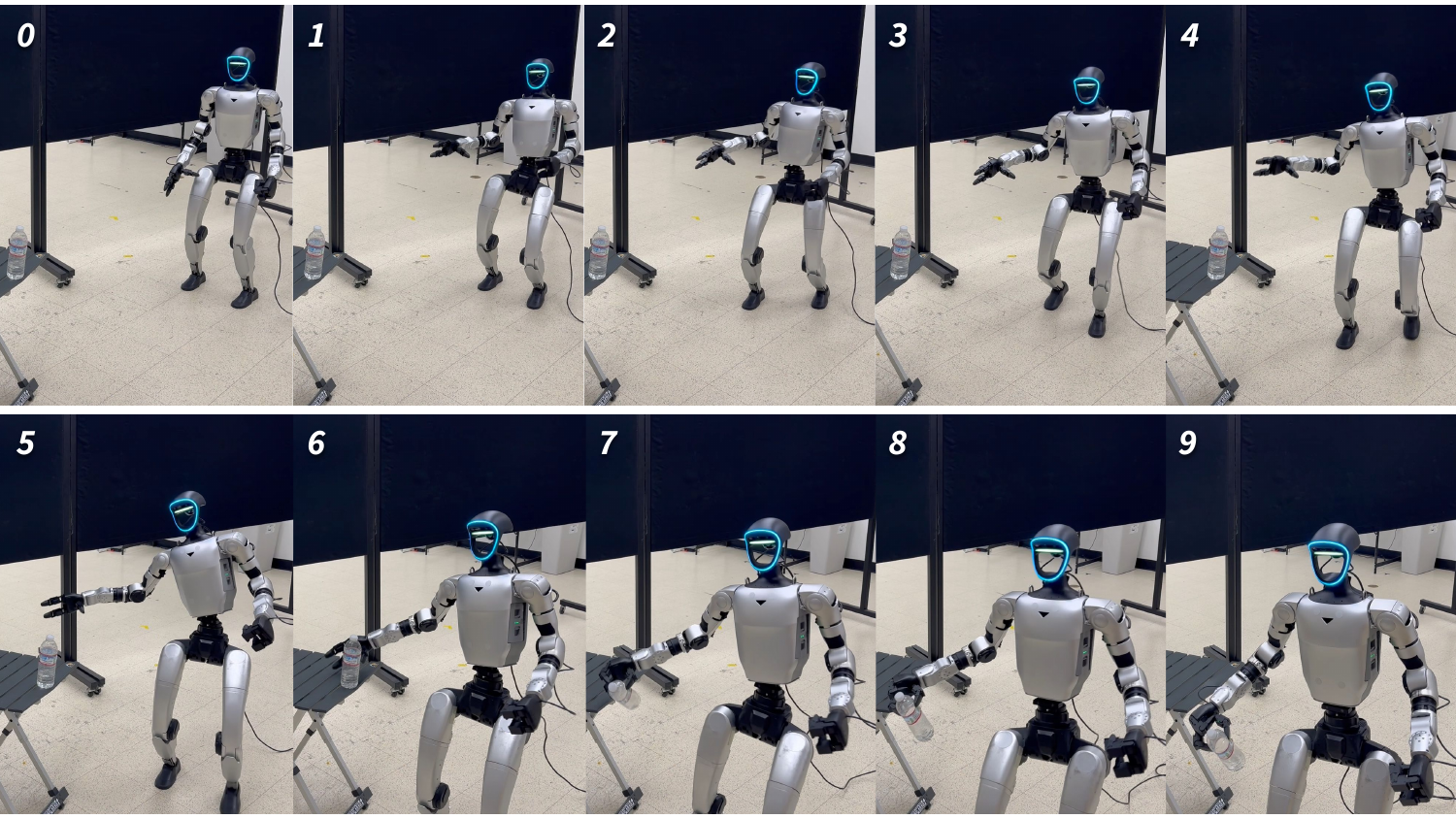}
    \caption{
    \textsc{WalkGrab} real-world demo.
    }
    \label{fig:walkgrab_supply}
\end{figure}

\begin{figure}[t]
    \centering
    \includegraphics[width=\linewidth]{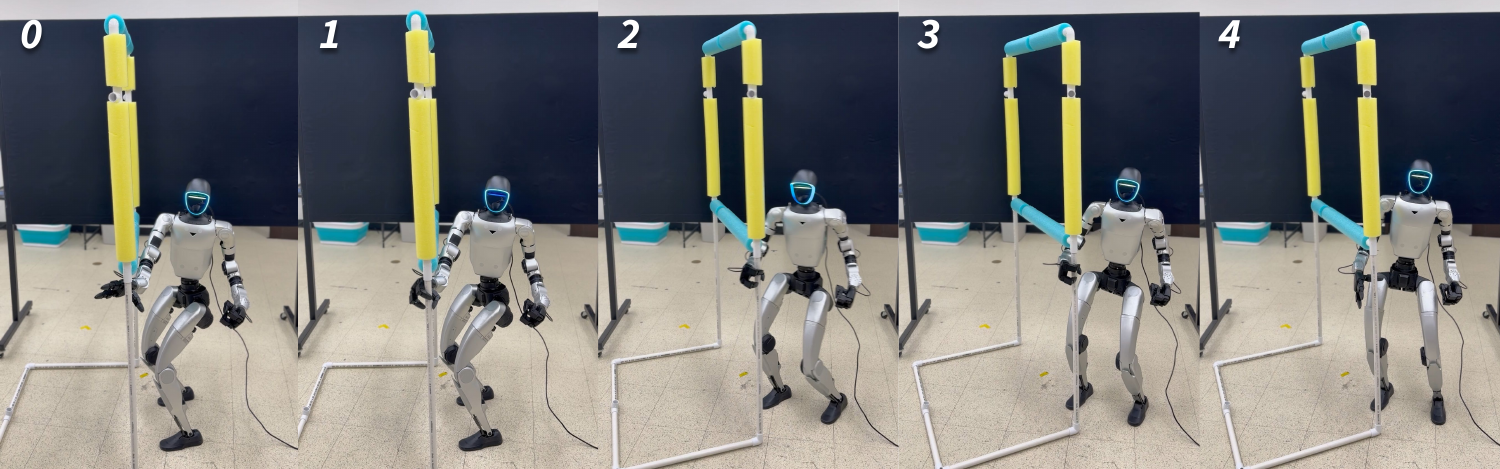}
    \caption{
    \textsc{OpenFridge} real-world demo. Due to facility constraints, we use a simplified mock-up instead of a full refrigerator door, focusing on the core behavior of maintaining a grasp while stepping backward to pull the object open.
    }
    \label{fig:openfridge_supply}
\end{figure}

This section provides additional qualitative hardware visualizations and clarifies the hardware variant used for real-robot replay. The quantitative simulation experiments in the main paper are conducted on a Unitree G1 humanoid equipped with a 20-DoF WUJI dexterous hand. In contrast, the physical robot available for our hardware visualization uses a Unitree G1 humanoid equipped with a Dex3-1~\cite{unitree_dex3} dexterous hand. The hardware results in this section should therefore be interpreted as a qualitative trajectory replay on a G1+Dex3-1 platform, rather than as the same G1+WUJI configuration used for the reported simulation success rates.

CoorDex supports such hand variants through its factorized body-hand prior design. The body prior is responsible for whole-body locomotion, reaching, and wrist placement, while the hand prior is specific to the dexterous hand morphology. When replacing the hand, the same pipeline can be instantiated by training a hand specific tracking teacher and distilling it into a hand-specific latent prior and decoder. In our implementation, we train an additional Dex3-1 hand tracking teacher and distill it into a Dex3-1 hand prior using the same procedure as the WUJI hand prior. The downstream residual interface remains structurally unchanged.

For the hardware visualization, we replay recorded joint position trajectories on the physical G1+Dex3-1 robot using the low level joint-position control interface. These trajectories are used to visualize whether the generated body-hand motions are kinematically compatible with the real hardware configuration. We include qualitative snapshots for each task, including \textsc{WalkGrab}, \textsc{OpenFridge}, and \textsc{WalkPickTurn}.

\bibliography{references}  

@article{Peng_2018,
   title={DeepMimic: example-guided deep reinforcement learning of physics-based character skills},
   volume={37},
   ISSN={1557-7368},
   url={http://dx.doi.org/10.1145/3197517.3201311},
   DOI={10.1145/3197517.3201311},
   number={4},
   journal={ACM Transactions on Graphics},
   publisher={Association for Computing Machinery (ACM)},
   author={Peng, Xue Bin and Abbeel, Pieter and Levine, Sergey and van de Panne, Michiel},
   year={2018},
   month=July, pages={1–14} }

@misc{liao2025beyondmimicmotiontrackingversatile,
      title={BeyondMimic: From Motion Tracking to Versatile Humanoid Control via Guided Diffusion}, 
      author={Qiayuan Liao and Takara E. Truong and Xiaoyu Huang and Yuman Gao and Guy Tevet and Koushil Sreenath and C. Karen Liu},
      year={2025},
      eprint={2508.08241},
      archivePrefix={arXiv},
      primaryClass={cs.RO},
      url={https://arxiv.org/abs/2508.08241}, 
}

@article{zhao2025dexh2r,
  title={Dexh2r: Task-oriented dexterous manipulation from human to robots},
  author={Zhao, Shuqi and Zhu, Xinghao and Chen, Yuxin and Li, Chenran and Xie, Yichen and Zhang, Xiang and Ding, Mingyu and Tomizuka, Masayoshi},
  journal={IEEE/ASME Transactions on Mechatronics},
  year={2025},
  publisher={IEEE}
}

@inproceedings{liang2025dexhanddiff,
  title={Dexhanddiff: Interaction-aware diffusion planning for adaptive dexterous manipulation},
  author={Liang, Zhixuan and Mu, Yao and Wang, Yixiao and Chen, Tianxing and Shao, Wenqi and Zhan, Wei and Tomizuka, Masayoshi and Luo, Ping and Ding, Mingyu},
  booktitle={Proceedings of the Computer Vision and Pattern Recognition Conference},
  pages={1745--1755},
  year={2025}
}

@inproceedings{zhang2026unidex,
  title={Unidex: A robot foundation suite for universal dexterous hand control from egocentric human videos},
  author={Zhang, Gu and Xu, Qicheng and Zhang, Haozhe and Ma, Jianhan and He, Long and Bao, Yiming and Ping, Zeyu and Yuan, Zhecheng and Lu, Chenhao and Yuan, Chengbo and others},
  booktitle={Proceedings of the IEEE/CVF Conference on Computer Vision and Pattern Recognition},
  pages={1841--1852},
  year={2026}
}

@misc{tan2026sphericallatentmotionprior,
      title={Spherical Latent Motion Prior for Physics-Based Simulated Humanoid Control}, 
      author={Jing Tan and Weisheng Xu and Xiangrui Jiang and Jiaxi Zhang and Kun Yang and Kai Wu and Jiaqi Xiong and Shiting Chen and Yangfan Li and Yixiao Feng and Yuetong Fang and Yujia Zou and Yiqun Song and Renjing Xu},
      year={2026},
      eprint={2603.01294},
      archivePrefix={arXiv},
      primaryClass={cs.RO},
      url={https://arxiv.org/abs/2603.01294}, 
}

@misc{xue2025openingsimtorealdoorhumanoid,
      title={Opening the Sim-to-Real Door for Humanoid Pixel-to-Action Policy Transfer}, 
      author={Haoru Xue and Tairan He and Zi Wang and Qingwei Ben and Wenli Xiao and Zhengyi Luo and Xingye Da and Fernando Castañeda and Guanya Shi and Shankar Sastry and Linxi "Jim" Fan and Yuke Zhu},
      year={2025},
      eprint={2512.01061},
      archivePrefix={arXiv},
      primaryClass={cs.RO},
      url={https://arxiv.org/abs/2512.01061}, 
}

@article{Peng_2022,
   title={ASE: large-scale reusable adversarial skill embeddings for physically simulated characters},
   volume={41},
   ISSN={1557-7368},
   url={http://dx.doi.org/10.1145/3528223.3530110},
   DOI={10.1145/3528223.3530110},
   number={4},
   journal={ACM Transactions on Graphics},
   publisher={Association for Computing Machinery (ACM)},
   author={Peng, Xue Bin and Guo, Yunrong and Halper, Lina and Levine, Sergey and Fidler, Sanja},
   year={2022},
   month=July, pages={1–17} }

@misc{schulman2017proximalpolicyoptimizationalgorithms,
      title={Proximal Policy Optimization Algorithms}, 
      author={John Schulman and Filip Wolski and Prafulla Dhariwal and Alec Radford and Oleg Klimov},
      year={2017},
      eprint={1707.06347},
      archivePrefix={arXiv},
      primaryClass={cs.LG},
      url={https://arxiv.org/abs/1707.06347}, 
}

@misc{unitree_g1,
  title        = {Unitree G1 Humanoid Robot},
  author       = {{Unitree Robotics}},
  year         = {2026},
  howpublished = {\url{https://www.unitree.com/g1}},
  note         = {Accessed: 2026-05-27}
}

@misc{wuji_hand,
  title        = {Wuji Hand Product Introduction},
  author       = {{WUJI TECH}},
  year         = {2026},
  howpublished = {\url{https://docs.wuji.tech/docs/en/wuji-hand/latest/overview/}},
  note         = {Accessed: 2026-05-27}
}

@misc{unitree_dex3,
  title        = {Unitree Dex3-1 Dexterous Hand},
  author       = {{Unitree Robotics}},
  year         = {2026},
  howpublished = {\url{https://www.unitree.com/Dex3-1}},
  note         = {Accessed: 2026-05-27}
}

@misc{he2024omnih2ouniversaldexteroushumantohumanoid,
      title={OmniH2O: Universal and Dexterous Human-to-Humanoid Whole-Body Teleoperation and Learning}, 
      author={Tairan He and Zhengyi Luo and Xialin He and Wenli Xiao and Chong Zhang and Weinan Zhang and Kris Kitani and Changliu Liu and Guanya Shi},
      year={2024},
      eprint={2406.08858},
      archivePrefix={arXiv},
      primaryClass={cs.RO},
      url={https://arxiv.org/abs/2406.08858}, 
}

@misc{liu2025opt2skillimitatingdynamicallyfeasiblewholebody,
      title={Opt2Skill: Imitating Dynamically-feasible Whole-Body Trajectories for Versatile Humanoid Loco-Manipulation}, 
      author={Fukang Liu and Zhaoyuan Gu and Yilin Cai and Ziyi Zhou and Hyunyoung Jung and Jaehwi Jang and Shijie Zhao and Sehoon Ha and Yue Chen and Danfei Xu and Ye Zhao},
      year={2025},
      eprint={2409.20514},
      archivePrefix={arXiv},
      primaryClass={cs.RO},
      url={https://arxiv.org/abs/2409.20514}, 
}

@misc{he2025hoverversatileneuralwholebody,
      title={HOVER: Versatile Neural Whole-Body Controller for Humanoid Robots}, 
      author={Tairan He and Wenli Xiao and Toru Lin and Zhengyi Luo and Zhenjia Xu and Zhenyu Jiang and Jan Kautz and Changliu Liu and Guanya Shi and Xiaolong Wang and Linxi Fan and Yuke Zhu},
      year={2025},
      eprint={2410.21229},
      archivePrefix={arXiv},
      primaryClass={cs.RO},
      url={https://arxiv.org/abs/2410.21229}, 
}

@article{homie2025,
  title = {HOMIE: Humanoid Loco-Manipulation with Isomorphic Exoskeleton Cockpit},
  author = {Ben, Qingwei and Jia, Feiyu and Zeng, Jia and Dong, Junting and Lin, Dahua and Pang, Jiangmiao},
  journal = {arXiv preprint arXiv:2502.13013},
  year = {2025},
  eprint = {2502.13013},
  archivePrefix = {arXiv},
  primaryClass = {cs.RO}
}

@misc{li2025maniptransefficientdexterousbimanual,
      title={ManipTrans: Efficient Dexterous Bimanual Manipulation Transfer via Residual Learning}, 
      author={Kailin Li and Puhao Li and Tengyu Liu and Yuyang Li and Siyuan Huang},
      year={2025},
      eprint={2503.21860},
      archivePrefix={arXiv},
      primaryClass={cs.RO},
      url={https://arxiv.org/abs/2503.21860}, 
}

@inproceedings{qin2023anyteleop,
  title     = {AnyTeleop: A General Vision-Based Dexterous Robot Arm-Hand Teleoperation System},
  author    = {Qin, Yuzhe and Yang, Wei and Huang, Binghao and Van Wyk, Karl and Su, Hao and Wang, Xiaolong and Chao, Yu-Wei and Fox, Dieter},
  booktitle = {Robotics: Science and Systems},
  year      = {2023}
}

@misc{zhao2026agilecomprehensiveworkflowhumanoid,
      title={AGILE: A Comprehensive Workflow for Humanoid Loco-Manipulation Learning}, 
      author={Huihua Zhao and Rafael Cathomen and Lionel Gulich and Wei Liu and Efe Arda Ongan and Michael Lin and Shalin Jain and Soha Pouya and Yan Chang},
      year={2026},
      eprint={2603.20147},
      archivePrefix={arXiv},
      primaryClass={cs.RO},
      url={https://arxiv.org/abs/2603.20147}, 
}

@article{viral2025,
  title = {VIRAL: Visual Sim-to-Real at Scale for Humanoid Loco-Manipulation},
  author = {He, Tairan and Wang, Zi and Xue, Haoru and Ben, Qingwei and Luo, Zhengyi and Xiao, Wenli and Yuan, Ye and Da, Xingye and Castaneda, Fernando and Sastry, Shankar and Liu, Changliu and Shi, Guanya and Fan, Linxi and Zhu, Yuke},
  journal = {arXiv preprint arXiv:2511.15200},
  year = {2025},
  eprint = {2511.15200},
  archivePrefix = {arXiv},
  primaryClass = {cs.RO}
}

@misc{jiang2025wholebodyvlaunifiedlatentvla,
      title={WholeBodyVLA: Towards Unified Latent VLA for Whole-Body Loco-Manipulation Control}, 
      author={Haoran Jiang and Jin Chen and Qingwen Bu and Li Chen and Modi Shi and Yanjie Zhang and Delong Li and Chuanzhe Suo and Chuang Wang and Zhihui Peng and Hongyang Li},
      year={2025},
      eprint={2512.11047},
      archivePrefix={arXiv},
      primaryClass={cs.RO},
      url={https://arxiv.org/abs/2512.11047}, 
}

@misc{he2024learninghumantohumanoidrealtimewholebody,
      title={Learning Human-to-Humanoid Real-Time Whole-Body Teleoperation}, 
      author={Tairan He and Zhengyi Luo and Wenli Xiao and Chong Zhang and Kris Kitani and Changliu Liu and Guanya Shi},
      year={2024},
      eprint={2403.04436},
      archivePrefix={arXiv},
      primaryClass={cs.RO},
      url={https://arxiv.org/abs/2403.04436}, 
}

@misc{fu2024humanplushumanoidshadowingimitation,
      title={HumanPlus: Humanoid Shadowing and Imitation from Humans}, 
      author={Zipeng Fu and Qingqing Zhao and Qi Wu and Gordon Wetzstein and Chelsea Finn},
      year={2024},
      eprint={2406.10454},
      archivePrefix={arXiv},
      primaryClass={cs.RO},
      url={https://arxiv.org/abs/2406.10454}, 
}

@misc{cheng2024expressivewholebodycontrolhumanoid,
      title={Expressive Whole-Body Control for Humanoid Robots}, 
      author={Xuxin Cheng and Yandong Ji and Junming Chen and Ruihan Yang and Ge Yang and Xiaolong Wang},
      year={2024},
      eprint={2402.16796},
      archivePrefix={arXiv},
      primaryClass={cs.RO},
      url={https://arxiv.org/abs/2402.16796}, 
}

@misc{ji2025exbody2advancedexpressivehumanoid,
      title={ExBody2: Advanced Expressive Humanoid Whole-Body Control}, 
      author={Mazeyu Ji and Xuanbin Peng and Fangchen Liu and Jialong Li and Ge Yang and Xuxin Cheng and Xiaolong Wang},
      year={2025},
      eprint={2412.13196},
      archivePrefix={arXiv},
      primaryClass={cs.RO},
      url={https://arxiv.org/abs/2412.13196}, 
}

@misc{zhang2025falconlearningforceadaptivehumanoid,
      title={FALCON: Learning Force-Adaptive Humanoid Loco-Manipulation}, 
      author={Yuanhang Zhang and Yifu Yuan and Prajwal Gurunath and Ishita Gupta and Shayegan Omidshafiei and Ali-akbar Agha-mohammadi and Marcell Vazquez-Chanlatte and Liam Pedersen and Tairan He and Guanya Shi},
      year={2025},
      eprint={2505.06776},
      archivePrefix={arXiv},
      primaryClass={cs.RO},
      url={https://arxiv.org/abs/2505.06776}, 
}

@misc{zhao2025resmimic,
  title={ResMimic: From General Motion Tracking to Humanoid Whole-body Loco-Manipulation via Residual Learning},
  author={Zhao, Siheng and Ze, Yanjie and Wang, Yue and Liu, C. Karen and Abbeel, Pieter and Shi, Guanya and Duan, Rocky},
  year={2025},
  eprint={2510.05070},
  archivePrefix={arXiv},
  primaryClass={cs.RO}
}

@misc{fu2025demohlmdemonstrationgeneralizablehumanoid,
      title={DemoHLM: From One Demonstration to Generalizable Humanoid Loco-Manipulation}, 
      author={Yuhui Fu and Feiyang Xie and Chaoyi Xu and Jing Xiong and Haoqi Yuan and Zongqing Lu},
      year={2025},
      eprint={2510.11258},
      archivePrefix={arXiv},
      primaryClass={cs.RO},
      url={https://arxiv.org/abs/2510.11258}, 
}

@misc{kuang2025skillblenderversatilehumanoidwholebody,
      title={SkillBlender: Towards Versatile Humanoid Whole-Body Loco-Manipulation via Skill Blending}, 
      author={Yuxuan Kuang and Haoran Geng and Amine Elhafsi and Tan-Dzung Do and Pieter Abbeel and Jitendra Malik and Marco Pavone and Yue Wang},
      year={2025},
      eprint={2506.09366},
      archivePrefix={arXiv},
      primaryClass={cs.RO},
      url={https://arxiv.org/abs/2506.09366}, 
}

@misc{sun2025ulc,
  title={ULC: A Unified and Fine-Grained Controller for Humanoid Loco-Manipulation},
  author={Sun, Wandong and Feng, Luying and Liu, Yang and Cao, Baoshi and Jin, Yaochu and Xie, Zongwu},
  year={2025},
  eprint={2507.06905},
  archivePrefix={arXiv},
  primaryClass={cs.RO}
}

@misc{heng2026humdex,
  title={HumDex: Humanoid Dexterous Manipulation Made Easy},
  author={Heng, Liang and Tang, Yihe and Xu, Jiajun and Bao, Henghui and Huang, Di and Wang, Yue},
  year={2026},
  eprint={2603.12260},
  archivePrefix={arXiv},
  primaryClass={cs.RO}
}

@article{Peng_2021,
   title={AMP: adversarial motion priors for stylized physics-based character control},
   volume={40},
   ISSN={1557-7368},
   url={http://dx.doi.org/10.1145/3450626.3459670},
   DOI={10.1145/3450626.3459670},
   number={4},
   journal={ACM Transactions on Graphics},
   publisher={Association for Computing Machinery (ACM)},
   author={Peng, Xue Bin and Ma, Ze and Abbeel, Pieter and Levine, Sergey and Kanazawa, Angjoo},
   year={2021},
   month=July, pages={1–20} }

@article{tessler2023calm,
  title={CALM: Conditional Adversarial Latent Models for Directable Virtual Characters},
  author={Tessler, Chen and Kasten, Yoni and Guo, Yunrong and Mannor, Shie and Chechik, Gal and Peng, Xue Bin},
  journal={ACM Transactions on Graphics},
  year={2023}
}

@inproceedings{luo2024pulse,
  title={Universal Humanoid Motion Representations for Physics-Based Control},
  author={Luo, Zhengyi and Cao, Jinkun and Merel, Josh and Winkler, Alexander and Huang, Jing and Kitani, Kris and Xu, Weipeng},
  booktitle={International Conference on Learning Representations},
  year={2024},
  eprint={2310.04582},
  archivePrefix={arXiv}
}

@misc{wang2023dexgraspnetlargescaleroboticdexterous,
      title={DexGraspNet: A Large-Scale Robotic Dexterous Grasp Dataset for General Objects Based on Simulation}, 
      author={Ruicheng Wang and Jialiang Zhang and Jiayi Chen and Yinzhen Xu and Puhao Li and Tengyu Liu and He Wang},
      year={2023},
      eprint={2210.02697},
      archivePrefix={arXiv},
      primaryClass={cs.RO},
      url={https://arxiv.org/abs/2210.02697}, 
}

@misc{li2023gendexgraspgeneralizabledexterousgrasping,
      title={GenDexGrasp: Generalizable Dexterous Grasping}, 
      author={Puhao Li and Tengyu Liu and Yuyang Li and Yiran Geng and Yixin Zhu and Yaodong Yang and Siyuan Huang},
      year={2023},
      eprint={2210.00722},
      archivePrefix={arXiv},
      primaryClass={cs.RO},
      url={https://arxiv.org/abs/2210.00722}, 
}

@misc{wei2025mathcaldrograspunifiedrepresentation,
      title={$\mathcal{D(R,O)}$ Grasp: A Unified Representation of Robot and Object Interaction for Cross-Embodiment Dexterous Grasping}, 
      author={Zhenyu Wei and Zhixuan Xu and Jingxiang Guo and Yiwen Hou and Chongkai Gao and Zhehao Cai and Jiayu Luo and Lin Shao},
      year={2025},
      eprint={2410.01702},
      archivePrefix={arXiv},
      primaryClass={cs.RO},
      url={https://arxiv.org/abs/2410.01702}, 
}

@misc{wei2026handruleallcanonical,
      title={One Hand to Rule Them All: Canonical Representations for Unified Dexterous Manipulation}, 
      author={Zhenyu Wei and Yunchao Yao and Mingyu Ding},
      year={2026},
      eprint={2602.16712},
      archivePrefix={arXiv},
      primaryClass={cs.RO},
      url={https://arxiv.org/abs/2602.16712}, 
}

@misc{zhan2024oakink2datasetbimanualhandsobject,
      title={OAKINK2: A Dataset of Bimanual Hands-Object Manipulation in Complex Task Completion}, 
      author={Xinyu Zhan and Lixin Yang and Yifei Zhao and Kangrui Mao and Hanlin Xu and Zenan Lin and Kailin Li and Cewu Lu},
      year={2024},
      eprint={2403.19417},
      archivePrefix={arXiv},
      primaryClass={cs.CV},
      url={https://arxiv.org/abs/2403.19417}, 
}

@misc{jiang2025dexmimicgenautomateddatageneration,
      title={DexMimicGen: Automated Data Generation for Bimanual Dexterous Manipulation via Imitation Learning}, 
      author={Zhenyu Jiang and Yuqi Xie and Kevin Lin and Zhenjia Xu and Weikang Wan and Ajay Mandlekar and Linxi Fan and Yuke Zhu},
      year={2025},
      eprint={2410.24185},
      archivePrefix={arXiv},
      primaryClass={cs.RO},
      url={https://arxiv.org/abs/2410.24185}, 
}

@misc{brüdigam2024jactaversatileplannerlearning,
      title={Jacta: A Versatile Planner for Learning Dexterous and Whole-body Manipulation}, 
      author={Jan Brüdigam and Ali-Adeeb Abbas and Maks Sorokin and Kuan Fang and Brandon Hung and Maya Guru and Stefan Sosnowski and Jiuguang Wang and Sandra Hirche and Simon Le Cleac'h},
      year={2024},
      eprint={2408.01258},
      archivePrefix={arXiv},
      primaryClass={cs.RO},
      url={https://arxiv.org/abs/2408.01258}, 
}

@article{mittal2025isaaclab,
  title={Isaac Lab: A GPU-Accelerated Simulation Framework for Multi-Modal Robot Learning},
  author={Mayank Mittal and Pascal Roth and James Tigue and Antoine Richard and Octi Zhang and Peter Du and Antonio Serrano-Muñoz and Xinjie Yao and René Zurbrügg and Nikita Rudin and Lukasz Wawrzyniak and Milad Rakhsha and Alain Denzler and Eric Heiden and Ales Borovicka and Ossama Ahmed and Iretiayo Akinola and Abrar Anwar and Mark T. Carlson and Ji Yuan Feng and Animesh Garg and Renato Gasoto and Lionel Gulich and Yijie Guo and M. Gussert and Alex Hansen and Mihir Kulkarni and Chenran Li and Wei Liu and Viktor Makoviychuk and Grzegorz Malczyk and Hammad Mazhar and Masoud Moghani and Adithyavairavan Murali and Michael Noseworthy and Alexander Poddubny and Nathan Ratliff and Welf Rehberg and Clemens Schwarke and Ritvik Singh and James Latham Smith and Bingjie Tang and Ruchik Thaker and Matthew Trepte and Karl Van Wyk and Fangzhou Yu and Alex Millane and Vikram Ramasamy and Remo Steiner and Sangeeta Subramanian and Clemens Volk and CY Chen and Neel Jawale and Ashwin Varghese Kuruttukulam and Michael A. Lin and Ajay Mandlekar and Karsten Patzwaldt and John Welsh and Huihua Zhao and Fatima Anes and Jean-Francois Lafleche and Nicolas Moënne-Loccoz and Soowan Park and Rob Stepinski and Dirk Van Gelder and Chris Amevor and Jan Carius and Jumyung Chang and Anka He Chen and Pablo de Heras Ciechomski and Gilles Daviet and Mohammad Mohajerani and Julia von Muralt and Viktor Reutskyy and Michael Sauter and Simon Schirm and Eric L. Shi and Pierre Terdiman and Kenny Vilella and Tobias Widmer and Gordon Yeoman and Tiffany Chen and Sergey Grizan and Cathy Li and Lotus Li and Connor Smith and Rafael Wiltz and Kostas Alexis and Yan Chang and David Chu and Linxi "Jim" Fan and Farbod Farshidian and Ankur Handa and Spencer Huang and Marco Hutter and Yashraj Narang and Soha Pouya and Shiwei Sheng and Yuke Zhu and Miles Macklin and Adam Moravanszky and Philipp Reist and Yunrong Guo and David Hoeller and Gavriel State},
  journal={arXiv preprint arXiv:2511.04831},
  year={2025},
  url={https://arxiv.org/abs/2511.04831}
}

@misc{xie2025kungfubotphysicsbasedhumanoidwholebody,
      title={KungfuBot: Physics-Based Humanoid Whole-Body Control for Learning Highly-Dynamic Skills}, 
      author={Weiji Xie and Jinrui Han and Jiakun Zheng and Huanyu Li and Xinzhe Liu and Jiyuan Shi and Weinan Zhang and Chenjia Bai and Xuelong Li},
      year={2025},
      eprint={2506.12851},
      archivePrefix={arXiv},
      primaryClass={cs.RO},
      url={https://arxiv.org/abs/2506.12851}, 
}

@misc{ze2025twistteleoperatedwholebodyimitation,
      title={TWIST: Teleoperated Whole-Body Imitation System}, 
      author={Yanjie Ze and Zixuan Chen and João Pedro Araújo and Zi-ang Cao and Xue Bin Peng and Jiajun Wu and C. Karen Liu},
      year={2025},
      eprint={2505.02833},
      archivePrefix={arXiv},
      primaryClass={cs.RO},
      url={https://arxiv.org/abs/2505.02833}, 
}

@misc{ze2025twist2scalableportableholistic,
      title={TWIST2: Scalable, Portable, and Holistic Humanoid Data Collection System}, 
      author={Yanjie Ze and Siheng Zhao and Weizhuo Wang and Angjoo Kanazawa and Rocky Duan and Pieter Abbeel and Guanya Shi and Jiajun Wu and C. Karen Liu},
      year={2025},
      eprint={2511.02832},
      archivePrefix={arXiv},
      primaryClass={cs.RO},
      url={https://arxiv.org/abs/2511.02832}, 
}

@misc{li2025amoadaptivemotionoptimization,
      title={AMO: Adaptive Motion Optimization for Hyper-Dexterous Humanoid Whole-Body Control}, 
      author={Jialong Li and Xuxin Cheng and Tianshu Huang and Shiqi Yang and Ri-Zhao Qiu and Xiaolong Wang},
      year={2025},
      eprint={2505.03738},
      archivePrefix={arXiv},
      primaryClass={cs.RO},
      url={https://arxiv.org/abs/2505.03738}, 
}

@misc{chen2025gmtgeneralmotiontracking,
      title={GMT: General Motion Tracking for Humanoid Whole-Body Control}, 
      author={Zixuan Chen and Mazeyu Ji and Xuxin Cheng and Xuanbin Peng and Xue Bin Peng and Xiaolong Wang},
      year={2025},
      eprint={2506.14770},
      archivePrefix={arXiv},
      primaryClass={cs.RO},
      url={https://arxiv.org/abs/2506.14770}, 
}

@misc{luo2026sonicsupersizingmotiontracking,
      title={SONIC: Supersizing Motion Tracking for Natural Humanoid Whole-Body Control}, 
      author={Zhengyi Luo and Ye Yuan and Tingwu Wang and Chenran Li and Fernando Castañeda and Sirui Chen and Zi-Ang Cao and Jiefeng Li and David Minor and Qingwei Ben and Jinhyung Park and David Sami and Zi Wang and Xingye Da and Runyu Ding and Cyrus Hogg and Lina Song and Edy Lim and Eugene Jeong and Tairan He and Haoru Xue and Wenli Xiao and Simon Yuen and Jan Kautz and Yan Chang and Umar Iqbal and Linxi "Jim" Fan and Yuke Zhu},
      year={2026},
      eprint={2511.07820},
      archivePrefix={arXiv},
      primaryClass={cs.RO},
      url={https://arxiv.org/abs/2511.07820}, 
}

\end{document}